\documentclass{article} 
\usepackage{iclr2024_conference,times}

\usepackage{amsmath,amsfonts,bm}

\def\eqref#1{equation~\ref{#1}}

\def\1{\bm{1}}

\DeclareMathAlphabet{\mathsfit}{\encodingdefault}{\sfdefault}{m}{sl}
\SetMathAlphabet{\mathsfit}{bold}{\encodingdefault}{\sfdefault}{bx}{n}

\newcommand{\sigmoid}{\sigma}

\usepackage{hyperref}
\usepackage{url}
\usepackage{graphicx}
\usepackage{booktabs}
\usepackage{multirow}
\usepackage[capitalise]{cleveref}

\title{Beyond the Benchmark: Detecting Diverse Anomalies in Videos}

\author{Yoav Arad, Michael Werman \\
Department of Computer Science\\
The Hebrew University of Jerusalem, Israel\\
Jerusalem, Israel \\
\texttt{\{yoav.arad,michael.werman\}@mail.huji.ac.il} \\
}

\def\ADset{HMDB-AD}
\def\VIOset{HMDB-Violence}
\def\methodShort{MFAD}
\def\methodLong{Multi-Frame Anomaly Detection}

\iclrfinalcopy 
\begin{document}
\maketitle

\begin{abstract}
Video Anomaly Detection (VAD) plays a crucial role in modern surveillance systems, aiming to  identify various anomalies in real-world situations. However, current benchmark datasets predominantly emphasize simple, single-frame anomalies such as novel object detection. This narrow focus restricts the advancement of VAD models. In this research, we advocate for an expansion of VAD investigations to encompass intricate anomalies that extend beyond conventional benchmark boundaries.
To facilitate this, we introduce two  datasets, \ADset\ and \VIOset, to challenge models with diverse action-based anomalies. These datasets are derived from the HMDB51 action recognition dataset.
We further present \methodLong\ (\methodShort), a novel method built upon the AI-VAD framework. 
AI-VAD utilizes single-frame features such as pose estimation and deep image encoding, and two-frame features such as object velocity. They then apply a density estimation algorithm to compute anomaly scores. To address complex multi-frame anomalies, we add a deep video encoding features capturing long-range temporal dependencies, and logistic regression  to enhance final score calculation.
Experimental results confirm our assumptions, highlighting existing models limitations with new anomaly types. \methodShort\ excels in both simple and complex anomaly detection scenarios. 
\end{abstract}

\section{Introduction}
\label{sec:intro}

As the volume of recorded video content continues to grow, the need for robust and efficient video anomaly detection methods increases. The ability to automatically identify unusual events or behaviors within videos not only holds the promise of enhancing security but also offers the potential to reduce the manpower required for monitoring. However, achieving truly effective video anomaly detection remains a significant unsolved challenge, due to the diverse range of anomalies that can occur in real-world scenarios.

By nature, anomalous behaviors are rare. Thus, video anomaly detection (VAD) is often treated as a semi-supervised problem, where models are trained exclusively on normal videos and must subsequently distinguish between normal and abnormal videos during inference.

While current benchmark datasets vary in complexity, they share a common limitation in their narrow definition of anomalies. The three datasets, UCSD Ped2 \citep{mahadevan_anomaly_2010}, CUHK Avenue \citep{lu_abnormal_2013}, and ShanghaiTech Campus \citep{liu_future_2018}, tend to limit anomalies primarily to novel object detection (Ped2, ShanghaiTech) or simple movement anomalies (Avenue). 

Recent advancements in video anomaly detection  predominantly relied on analyzing a few frames or even individual frames in isolation. Researchers predominantly choose between two approaches: reconstruction-based and prediction-based methods. Reconstruction-based methods \citep{nguyen_anomaly_2019, luo_remembering_2017, gong_memorizing_2019, fan_video_2018, hasan_learning_2016} typically employ auto-encoders to learn representations of normal frames, reconstructing them accurately, while anomalous frames result in a higher reconstruction error. Prediction-based methods \citep{liu_future_2018, lu_future_2019, yu_cloze_2020, park_learning_2020} focus on predicting the next frame from a sequence of consecutive frames.

These few-frame based methods achieved impressive results, surpassing an AUC score of 99\% \citep{reiss_attribute-based_2022, liu_hybrid_2021} on Ped2, over 93\% \citep{reiss_attribute-based_2022} on Avenue, and exceeding 85\% \citep{reiss_attribute-based_2022, barbalau_ssmtl_2023} on ShanghaiTech, the most complex of the benchmark datasets.

Without a shift in research focus and assumptions, the existing datasets, results, and recurring research patterns may suggest that the field of video anomaly detection is nearing a plateau.

This paper emphasizes the necessity of broadening the scope of what constitutes an anomaly. We propose two novel datasets specifically designed to assess the detection of complex action-based anomalies. These datasets, referred to as \ADset\ and \VIOset, build upon the HMDB51 action recognition dataset and define different actions as normal or abnormal activities. By analyzing the performance of various methods on our datasets, we underscore the limitations of existing approaches and advocate for further research on more comprehensive anomaly types.

Building upon the foundation laid by AI-VAD \citep{reiss_attribute-based_2022}, we introduce \methodLong\ (\methodShort), a novel method aimed at achieving balanced performance, excelling in both simple and complex anomaly detection. AI-VAD utilizes a two-step approach: first, it extracts multiple features  and then employs density estimation algorithms to calculate anomaly scores. 
In their work, they rely on single-frame features like deep image encoding (using a pretrained encoder) and human pose estimations, along with two-frame features such as object velocity. We extend this method by introducing deep video encoding features to capture multi-frame, long-range temporal relationships. \methodShort\ adheres to the AI-VAD framework, computing final scores for each feature using a density estimation algorithm. Additionally, we incorporate  logistic regression  to enhance the relationships between different feature scores and achieve more accurate final scores.

We extensively evaluate our method on classic benchmark datasets as well as on our newly proposed datasets. The experiments validate the added value of both video encoding features and the logistic regression module. Our method achieves competitive results on Ped2, Avenue, and ShanghaiTech, and greatly outperforms recent methods on \ADset\ and \VIOset. As a result, it offers a more versatile video anomaly detection solution capable of detecting a broader range of anomalies across various scenarios.

Our key contributions are:
\begin{itemize}
    \item We highlight the limitations of current video anomaly detection benchmarks and advocate for further research in general video anomaly detection.
    \item We present \methodShort, a novel method capable of effectively handling both simple, few-frame anomalies and complex, multi-frame anomalies.
    \item We provide two datasets designed for assessing a model's performance on multi-frame action-based anomalies.
\end{itemize}

\section{Related Work}
\label{sec:related_work}

\subsection{Video Anomaly Detection Datasets}
The datasets commonly used in video anomaly detection can be broadly categorized into two groups, reflecting the shift brought about by the advent of deep learning from approximately 2013 to 2018. Detailed comparison can be found in \cref{tab:vad_datasets}

Early datasets are notably smaller and often considered practically solved, include Subway Entrance \citep{adam_robust_2008}, Subway Exit \citep{adam_robust_2008}, UMN \citep{university_of_minnesota_unusual_2006}, UCSD Ped1 \citep{mahadevan_anomaly_2010}, UCSD Ped2 \citep{mahadevan_anomaly_2010}, and CUHK Avenue \citep{lu_abnormal_2013}. With the exception of UMN, these datasets feature only a single scene.

In contrast, more recent datasets have grown significantly in both scale and complexity. This newer group includes ShanghaiTech Campus \citep{liu_future_2018}, Street Scene \citep{ramachandra_street_2020}, IITB Corridor \citep{rodrigues_multi-timescale_2020}, UBNormal \citep{acsintoae_ubnormal_2022}, and the most recent and largest of them all, NWPU Campus \citep{cao_new_2023}. 

It's worth noting that among these datasets, only three have gained popularity as benchmarks: UCSD Ped2, CUHK Avenue, and ShanghaiTech Campus. However, as discussed in this paper, each of these benchmarks has its own set of limitations that motivate the need for further research in the field of video anomaly detection.

\begin{table}[ht]
  \caption{Comparison of the different video anomaly detection datasets.}
  \label{tab:vad_datasets}
  \begin{center}
  \begin{tabular}{@{}lccc@{}}
    \toprule
    \multicolumn{1}{c}{\multirow{2}{*}{Dataset}} & \multicolumn{3}{c}{\# Frames} \\
    & Total & Train & Test \\
    \midrule
    Subway Entrance \citep{adam_robust_2008} & 86,535 & 18,000 & 68,535 \\
    Subway Exit \citep{adam_robust_2008} & 38,940 & 4,500 & 34,440 \\
    UMN \citep{university_of_minnesota_unusual_2006} & 7,741 & -- & -- \\
    USCD Ped1 \citep{mahadevan_anomaly_2010} & 14,000 & 6,800 & 7,200 \\
    USCD Ped2 \citep{mahadevan_anomaly_2010} & 4,560 & 2,550 & 2,010 \\
    CUHK Avenue \citep{lu_abnormal_2013} & 30,652 & 15,328 & 15,324 \\
    ShanghaiTech Campus \citep{liu_future_2018} & 317,398 & 274,515 & 42,883 \\
    Street Scene \citep{ramachandra_street_2020} & 203,257 & 56,847 & 146,410 \\
    IITB Corridor \citep{rodrigues_multi-timescale_2020} & 483,566 & 301,999 & 181,567 \\
    UBnormal \citep{acsintoae_ubnormal_2022} & 236,902 & 116,087 & 92,640 \\
    NWPU Campus \citep{cao_new_2023} & 1,466,073 & 1,082,014 & 384,059 \\
    \midrule
    \ADset\ (ours) & 92,585 & 58,790 & 33,795 \\
    \VIOset\ (ours) & 204,471 & 140,377 & 64,094 \\
    \bottomrule
  \end{tabular}
  \end{center}
\end{table}

\subsection{HMDB51 Action Recognition Dataset}
The HMDB51 \citep{kuehne_hmdb_2011} dataset, originally designed for action recognition (AR), is relatively small in scale. It is a collection of 6,766 video clips distributed across 51 distinct categories. Most other datasets are significantly larger and more diverse: SSv2 \citep{qualcomm_moving_2018}, Kinetics-400 \citep{kay_kinetics_2017}, Kinetics-600 \citep{carreira_short_2018}, Kinetics-700-2020 \citep{smaira_short_2020} each consist of hundred of thousands of frames and hundreds of different classes. A comparison can be found in \cref{tab:ar_datasets}. 

The HMDB51 dataset draws content from various sources, ensuring diversity. In this dataset, each class consists of no less than 101 video clips.

\begin{table}[ht]
  \caption{Comparison of selected action recognition datasets.}
  \label{tab:ar_datasets}
  \begin{center}
  \begin{tabular}{@{}lcc@{}}
    \toprule
    \multicolumn{1}{c}{Dataset} & Total Clips & \# Classes \\
    \midrule
    UCF-101 \citep{soomro_ucf101_2012} & 13,320 & 101 \\
    ActivityNet-200 \citep{caba_heilbron_activitynet_2015} & 28,108 & 200 \\
    Something-Something v1 \citep{goyal_something_2017} & 108,499 & 174 \\
    Something-Something v2 \citep{qualcomm_moving_2018} & 220,847 & 174 \\
    Kinetics-400 \citep{kay_kinetics_2017} & 306,245 & 400 \\
    Kinetics-600 \citep{carreira_short_2018} & 495,547 & 600 \\
    Kinetics-700-2020 \citep{smaira_short_2020} & 647,907 & 700 \\
    \midrule
    HMDB51 \citep{kuehne_hmdb_2011} & 6,766 & 51 \\
    \bottomrule
  \end{tabular}
  \end{center}
\end{table}

\subsection{Video Anomaly Detection Methods}
\paragraph{Hand-crafted feature based methods}
Numerous methods, spanning both classical and contemporary approaches, adhere to a two-stage anomaly detection framework. This framework involves an initial step of extracting hand-crafted features, specifically selected by the researcher and not learned through a deep neural network model. Subsequently, another algorithm is applied to compute anomaly scores.

Early techniques used classic image and video features, including the histogram of oriented optical flow (HOF) \citep{chaudhry_histograms_2009, pers_histograms_2010, colque_histograms_2017, sabzalian_deep_2019}, histogram of oriented gradients (HOG) \citep{sabzalian_deep_2019}, and SIFT descriptors \citep{lowe_distinctive_2004}. 
In more recent developments, the proliferation of deep learning has facilitated the adoption of off-the-shelf models, such as object detectors, for feature extraction. For instance, in the case of AI-VAD \citep{reiss_attribute-based_2022}, a combination of pose estimations, optical flow predictions, object detection, and deep image encodings is used to construct robust feature representations.

Following feature extraction, classical methodologies often employed scoring techniques such as density estimation algorithms \citep{latecki_outlier_2007,glodek_ensemble_2013,eskin_geometric_2002}. Recent approaches have demonstrated the effectiveness of integrating these features with another learning model \citep{liu_hybrid_2021}.

\paragraph{Reconstruction and prediction based methods}
In recent years, the increasing prominence of deep learning has driven the widespread adoption of both reconstruction and prediction based methods in video anomaly detection.

Reconstruction-based \citep{nguyen_anomaly_2019, luo_remembering_2017, gong_memorizing_2019, fan_video_2018, hasan_learning_2016} approaches often utilize auto-encoders to learn representations of normal video frames and subsequently detect abnormal frames by identifying higher reconstruction errors. However, the powerful generalization ability of modern auto-encoders can often also reconstruct anomalies. Thus, making it harder to differentiate normal and abnormal frames.

Prediction-based  \citep{liu_future_2018, lu_future_2019, yu_cloze_2020, park_learning_2020} models forecast the subsequent frame by leveraging a sequence of preceding frames, employing time sensitive architectures such as LSTMs, memory networks, 3D auto-encoders and transformers. This predictive approach often yields superior results compared to similar reconstruction-based techniques \citep{park_learning_2020}, as it captures more complex forms of anomalies. Nevertheless, with the minimal differences between consecutive video frames, these methods face similar challenges to reconstruction-based approaches in respect to modern generators.

\paragraph{Auxiliary tasks methods}
Expanding beyond reconstruction and prediction, some models incorporate diverse self-supervised auxiliary tasks, with task success determining frame anomaly scores. These tasks include jigsaw puzzles \citep{wang_video_2022}, time direction detection \citep{wei_learning_2018}, rotation prediction \citep{gidaris_unsupervised_2018}, and more. 
\citet{barbalau_ssmtl_2023,georgescu_anomaly_2021}, train a single deep backbone on multiple self-supervised tasks and achieve state-of-the-art results on the benchmark datasets.

\section{Proposed Datasets}
\label{sec:datasets}
We introduce two novel datasets designed to assess the capability of various models in detecting forms of anomalies not covered by existing benchmarks. These datasets emphasize action-based anomalies, a category absent in current benchmarks. The first dataset, referred to as \ADset, aligns with the conventional definition of normal activities (walking and running) but challenges models with abnormal behaviors that demand a broader context for detection (climbing and performing a cartwheel). 
In contrast, the larger and more intricate \VIOset\ dataset divides 16 action categories into 7 violent (abnormal) and 9 non-violent (normal) activities. This categorization necessitates models to consider a wide range of behaviors when classifying events as either normal or abnormal, making it a closer representation of real-world scenarios. A comparison to other video anomaly detection datasets can be found in \cref{tab:vad_datasets}.

\paragraph{\ADset\ dataset}
\ADset\ is the simpler dataset among the two introduced in this paper. It consists of 995 video clips, divided into 680 training videos and 315 testing videos. Normal activities within this dataset are running and walking, aligning with their respective HMDB51 classes. Abnormal activities are climbing and performing a cartwheel. The training dataset  contains only of normal videos: 207 running videos and 473 walking videos. Meanwhile, the test dataset has both abnormal videos and randomly selected normal videos;  107 cartwheel videos, 108 climbing videos, 25 running videos, and 75 walking videos. Frames from the  videos can be viewed in \cref{app:ad_examples}.

\paragraph{\VIOset\ dataset}
\VIOset\ is the larger and more complex of the two datasets presented in this paper. It has 2,566 videos, with a distribution of 1,601 training videos and 965 testing videos. The train set has nine normal categories: running (221 videos), walking (517), waving (98), climbing (104), hugging (110), throwing (96), sitting (134), turning (222), and performing a cartwheel (99). In the test set, there are seven abnormal categories: falling (136), fencing (116), hitting (127), punching (126), using a sword (127), shooting (103), and kicking (130). Additionally, the test set includes 100 videos randomly sampled from the various normal categories: turning (18), walking (31), running (11), sitting (8), hugging (8), performing a cartwheel (8), climbing (4), throwing (6), and waving (6). The abnormal activities in HMDB-Violence are characterized by their violent nature. Examples can be viewed in \cref{app:vio_examples}.

\paragraph{Annotations}
We maintain a consistent labeling for every frame within a video. If a video represents a normal action category, all its frames are labeled as  normal. Conversely, if it belongs to an abnormal action category, all frames are marked as as abnormal. This simple labeling approach works, as the actions within these videos effectively occupy the entire duration, leaving minimal room for unrelated "spare" frames.

\section{\methodShort: \methodLong}
\label{sec:method}
Our method, \methodShort, consists of three key stages: feature extraction, per-feature score computation, and logistic regression. We extract four types of features: object velocities, human pose estimations, deep image encodings, and deep video encodings. For each of these features, we independently calculate density scores. We then employ a logistic regression model to  optimally   fuse the scores across these four feature kinds. Lastly, we  smooth, Gaussian, to produce the final anomaly scores.

\begin{figure}[ht]
\begin{center}
  \includegraphics[width=1\linewidth]{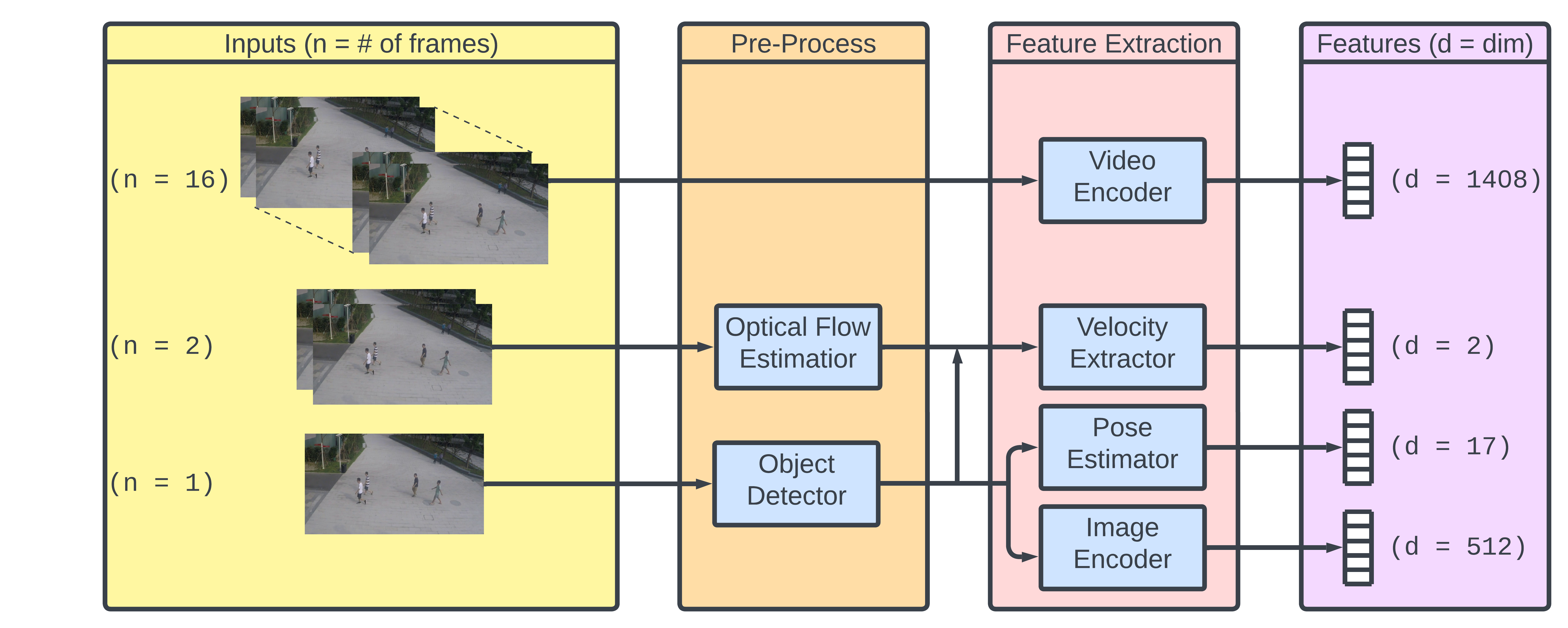}
  \end{center}
  \caption{An overview of our feature extraction process.}
  \label{fig:feature_extraction}
\end{figure}

\subsection{Feature Extraction}
\paragraph{Few-Frame Features}
In line with \citet{reiss_attribute-based_2022, liu_hybrid_2021}, we extract object bounding boxes and optical flows from each frame. We then extract human pose estimations, object velocities, and deep image encodings. These features are derived from individual frames (pose and image encoding) or pairs of frames (velocity) enabling the detection of straightforward anomalies such as novel objects.

\paragraph{Multi-Frame Features}
Recognizing the necessity for detecting complex anomalies that span multiple frames, we introduce a deep video encoder. This encoder captures features in a manner similar to deep image encoding but takes into account longer frame sequences (in our case, 16 frames). For this purpose, we leverage VideoMAEv2 \citep{wang_videomae_2023}, a state of the art video foundation model. Subsequently, we process these features in a fashion similar to AI-VAD \citep{reiss_attribute-based_2022}.

\subsection{Density Score Calculation}
We employ a Gaussian Mixture Model (GMM) for the two-dimensional velocity features and the k-nearest neighbors (kNN) algorithm for the high-dimensional pose, image encoding, and video encoding features. Subsequently, we compute the minimum and maximum density scores for the training set and use them to calibrate the test scores during inference.

\paragraph{Max Feature}
We add a fifth feature, denoted as $\max \in [0,1]^{\# frames}$. After calculating the density scores per feature, we aggregate them to a new feature that holds the maximum feature score per frame.
\begin{center}
    $\max = \max\{\text{pose score, velocity score, image encoding score, video encoding score}\}$
\end{center}
Our experiments show the added value of this feature.

\subsection{Logistic Regression}
In order to improve the accuracy of our final anomaly score computation, we incorporate logistic regression as the final step of our method.
In this setup, we denote $X\in [0,1]^{\# frames \times \# features}$ as the feature matrix and $y\in \{0,1\}^{\# frames}$ as our ground truth labels. Our final scores are:
\begin{center}
    $h_\theta(X) = \sigmoid(WX + B)$
\end{center}
where $\sigmoid(t) = \frac{1}{1+e^{-t}}$ is the sigmoid function and $\theta=(W,B)$ are the parameters we want to optimize. Our loss function is:
\begin{center}
    $L(h_\theta(X),y) = -y\log(h_\theta(X))-(1-y)\log(1-h_\theta(X))$    
\end{center}

During its training phase, we randomly sample a small fraction of the test frames for model training, while the remainder is used for evaluation. It's crucial to emphasize that the frames utilized for training are excluded from the evaluation process for our reported results, ensuring the validity of our findings.

The final step in our method is applying Gaussian smoothing to the anomaly scores.

\begin{figure}[ht]
\begin{center}
  \includegraphics[width=1\linewidth]{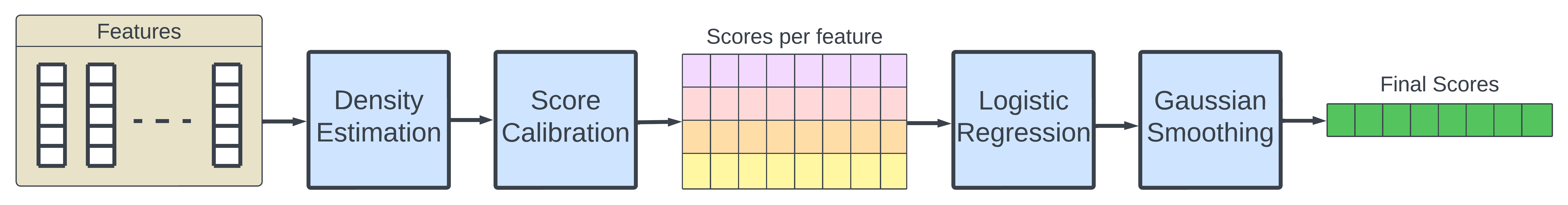}
  \end{center}
  \caption{An overview of our anomaly score calculation during inference.}
  \label{fig:score_calculation}
\end{figure}

\section{Experiments}
\label{sec:experiments}

\subsection{Datasets}
In addition to \ADset\ and \VIOset, we evaluate \methodShort\ on the three benchmark video anomaly detection datasets: UCSD Ped2, CUHK Avenue, and ShanghaiTech Campus. These datasets are primarily  outdoor surveillance camera footage, with the sole normal activity being pedestrian movement. Details are presented in \cref{tab:vad_datasets}.

\paragraph{UCSD Ped2}
The UCSD Ped2 dataset has 16 training videos and 12 testing videos, all situated within a single scene. Abnormal events in this dataset include the appearance of skateboards, bicycles, or cars within the video frame. Videos are standardized to a resolution of 240 × 360 pixels.

\paragraph{CUHK Avenue}
The CUHK Avenue dataset has 16 training videos and 21 testing videos, all within a single scene. Anomalies within this dataset are activities such as running, throwing objects, and bike riding. All videos have a resolution of 360 × 640 pixels.

\paragraph{ShanghaiTech Campus}
ShanghaiTech Campus stands as the largest and most complex dataset among the three, featuring 330 training videos and 107 testing videos distributed across 13 distinct scenes. Notably, two of these scenes involve non-stationary cameras, resulting in varying angles between videos of the same scene. Abnormal events primarily include running and the presence of cars and bikes. All videos have a resolution of 480 × 856 pixels.

\subsection{Implementation details}
\label{sec:implementation_details}
We adopt the code from AI-VAD for extracting velocity, pose, and deep image encoding features. For our new deep video features, we leverage the state-of-the-art video foundation model, VideoMAEv2 \citep{wang_videomae_2023}, with the publicly available pretrained weights, fine-tuned on the SSv2 dataset (\textit{vit\_g\_hybrid\_pt\_1200e\_ssv2\_ft}). 
Our encoding process is carried out on non-overlapping consecutive blocks of 16 frames, extracting Temporal Action Detection (TAD) features for each block. In our experiments we found no difference in results between non-overlapping blocks and sliding-window blocks.
When employing the nearest neighbors algorithm to the video encoding features, we set $k=1$.

Our code and a setup guide are available on \url{https://github.com/yoavarad/MFAD}.
\subsection{Anomaly Detection Results}
Our results are based on our optimal model configuration, see \cref{sec:ablation}. This configuration involves leveraging all four feature types and the $\max$ feature, while training a logistic regression model on a random 2\% of the test set frames for computing final anomaly scores. It's crucial to note that the data used for training the logistic regression model is not included in the evaluation. To ensure reliability, we repeat this final step 100 times and report the mean AUC result along with the standard deviation. The consistently low standard deviation across all datasets underscores the stability of our method.

\methodShort\ demonstrates competitive results on the well-established benchmark datasets, with modest differences of approximately -0.3\%, -0.8\%, and -1.1\% from the state-of-the-art results on Ped2, Avenue, and ShanghaiTech, respectively. The true strength of our approach becomes evident when applied to the newly introduced datasets, \ADset\ and \VIOset. On these datasets, we achieve substantial improvements of 19.9\% and 9.7\%, respectively. \methodShort\ was tested against four different methods on these new datasets, including AI-VAD \citep{reiss_attribute-based_2022}, upon which our work is built and is the state-of-the-art on the ShanghaiTech dataset. This substantial enhancement highlights the generalizability of our approach to various complex anomalies, without majorly impacting our detection ability of simple anomalies, underscoring the significance of our contributions. For detailed comparison see \cref{tab:vad_results}.

We further report the configuration of \methodShort\ without image encoding (IE) features, improving results on ShanghaiTech by 0.2\%.

\begin{table}[ht]
  \caption{Comparison to frame-level AUC. Best (bold), second (underlined) and third (italic) best results are highlighted. Image encoding features is denoted by IE.}
  \label{tab:vad_results}
  \begin{center}
  {\scriptsize \begin{tabular}{@{}lccccc@{}}
    \toprule
    Method                                         & Ped2               & Avenue             & ShanghaiTech                   & \ADset                     & \VIOset \\
    \midrule
    HF2-VAD \citep{liu_hybrid_2021}                & \textbf{99.3\%}    & 91.1\%             & 76.2\%                         & --                         & -- \\
    AED \citep{georgescu_background-agnostic_2022} & 98.7\%             & 92.3\%             & 82.7\%                         & --                         & -- \\
    HSC-VAD \citep{sun_hierarchical_2023}          & 98.1\%             & \textbf{93.7\%}    & 83.4\%                         & --                         & -- \\
    DLAN-AC \citep{yang_dynamic_2022}              & 97.6\%             & 89.9\%             & 74.7\%                         & --                         & -- \\
    SSMTL \citep{georgescu_anomaly_2021}           & 97.5\%             & 91.5\%             & 82.4\%                         & --                         & -- \\
    LBR-SPR \citep{yu_deep_2022}                   & 97.2\%             & 90.7\%             & 72.6\%                         & --                         & -- \\
    AMMCNet \citep{cai_appearance-motion_2021}     & 96.6\%             & 86.6\%             & 73.7\%                         & --                         & -- \\
    \midrule
    AI-VAD \citep{reiss_attribute-based_2022}      & \underline{99.1\%} & \underline{93.3\%} & \textbf{85.9\%}                & \textit{70.1\%}                     & \textit{70.5\%} \\
    Jigsaw Puzzles \citep{wang_video_2022}         & 99.0\%             & 92.2\%             & 84.3\%                         & 53.8\%                     & 52.7\% \\
    MNAD \citep{park_learning_2020}                & 97.0\%             & 88.5\%             & 70.5\%                         & 56.3\%                     & 51.3\% \\
    MPN \citep{lv_learning_2021}                   & 96.9\%             & 89.5\%             & 73.8\%                         & 58.8\%                     & 53.7\% \\
    \midrule
    \methodShort\ (Ours)              & \textit{99.0\% ± 0.5\%}     & \textit{92.9\% ± 0.5\%}     & \textit{84.8\% ± 0.4\%}               & \textbf{90.0\% ± 0.4\%}    & \textbf{80.2\% ± 0.2\%} \\
    \methodShort\ w/o IE (Ours)       & 98.4\% ± 0.7\%     & 90.7\% ± 0.5\%     & \underline{85.0\% ± 0.4\%}   & \underline{86.9\% ± 0.5\%}             & \underline{76.0\% ± 0.2\%} \\
    \bottomrule
  \end{tabular}}
  \end{center}
\end{table}

\begin{figure}[ht]
\begin{center}
  \includegraphics[width=0.24\linewidth]{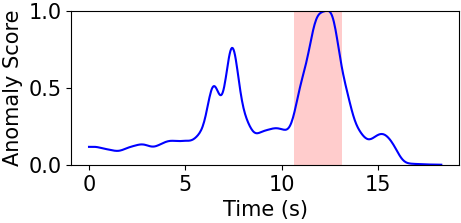}
  \includegraphics[width=0.24\linewidth]{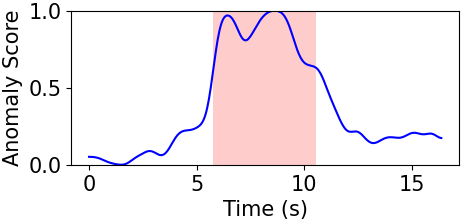}
  \includegraphics[width=0.24\linewidth]{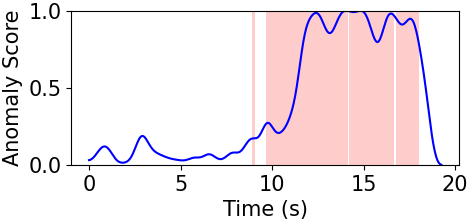}
  \includegraphics[width=0.24\linewidth]{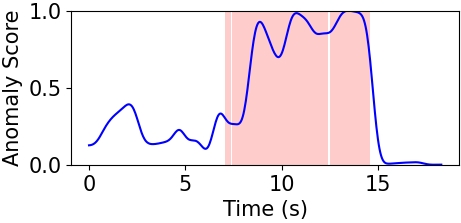}
  \includegraphics[width=0.24\linewidth]{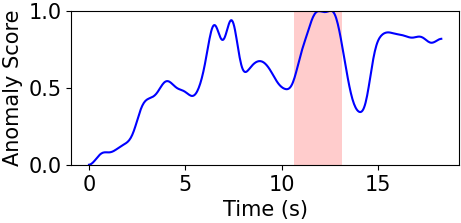}
  \includegraphics[width=0.24\linewidth]{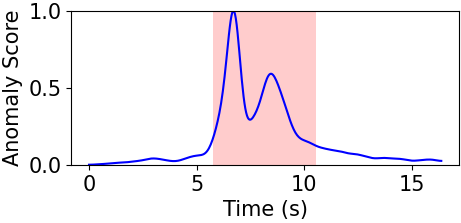}
  \includegraphics[width=0.24\linewidth]{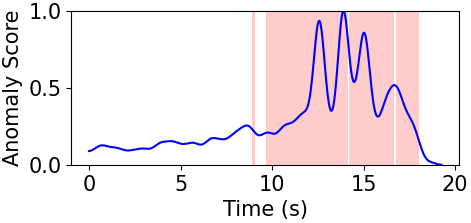}
  \includegraphics[width=0.24\linewidth]{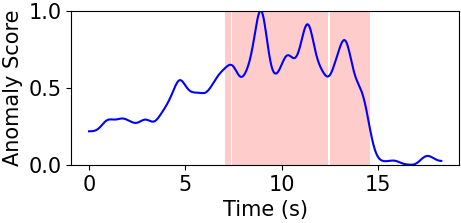}
  \end{center}
  \caption{Qualitative results from three ShanghaiTech videos: 01\_0028 (left), 03\_0032 ($2^{nd}$ left), 03\_0039 ($2^{nd}$ right), 07\_0008 (right). In each pair, \methodShort\ (top) is compared to AI-VAD \citep{reiss_attribute-based_2022} (bottom). Anomalous sections are highlighted in red, while the anomaly scores, ranging from 0 to 1, are the blue line. These videos feature complex, behavior-based anomalies rather than novel object detection scenarios, that are more common in this dataset. Clearly, \methodShort\ improves both detecting anomalies and accurately assessing normal parts of the video. Best viewed in color.}
  \label{fig:qual_results}
\end{figure}

In addition to quantitative evaluations, we conducted qualitative analyses on videos from the ShanghaiTech dataset, which feature more complex anomalies beyond novel object detection. These anomalies are shown in \cref{app:qual_examples}. The positive impact of our method is clearly evident in \cref{fig:qual_results}, where abnormal frames receive higher anomaly scores, while normal frames receive lower anomaly scores, further validating our method.

\subsection{Ablation Study}
\label{sec:ablation}
We perform an ablation study to determine two factors: the added benefit of the video encoding feature, and the most favorable configuration for the logistic regression module.

\paragraph{Feature Selection}
In their ablation study, \citet{reiss_attribute-based_2022} demonstrated the incremental value of their three distinct feature types: pose estimation, deep image encoding, and velocity features, as well as the added effect of Gaussian smoothing. Expanding upon their work, we test the impact of incorporating deep video encoding features in different forms. Our study consists of tests involving both video encoding features in isolation, the combination of all four feature types, and the addition of an extra $\max$ feature, that has the value of the maximum feature score per each frame. Furthermore, we explore the substitution of image encoding features with video encoding features due to their semantic similarity.

As presented in \cref{tab:feature_ablation}, utilizing solely video encoding features yields impressive performance for multi-frame anomalies. However, this specialization comes at the cost of lower performance on the traditional video anomaly detection datasets. On the other hand, employing all four feature types results in a comprehensive and well-balanced model that performs admirably across all datasets, even though it may not achieve the top rank in any specific dataset.

\begin{table}[ht]
  \caption{Comparison of different model configurations, evaluating the impact of various feature types, including pose features (P), velocity features (V), image encoding features (IE), and video encoding features (VE), on the model's performance. $\max$ is the max value between \{P, V, IE, VE\}. Best and second best results are in bold and underlined, respectively.}
  \label{tab:feature_ablation}
  \begin{center}
  {\small \begin{tabular}{@{}lccccc@{}}
    \toprule
    Configuration                                 & Ped2                & Avenue                   & ShanghaiTech                & \ADset             & \VIOset \\
    \midrule
    VE                                            & 80.3\%              & 87.9\%                   & 71.3\%                      & \underline{84.9\%} & \underline{75.8}\% \\
    P + V \citep{reiss_attribute-based_2022}      & \underline{98.7\%}  & 86.8\%                  & \textbf{85.9\%}              & 54.2\%             & 56.1\% \\
    P + V + IE \citep{reiss_attribute-based_2022} & \textbf{99.1\%}     & \textbf{93.5\%}         & \underline{85.1\%}           & 71.2\%             & 67.9\% \\
    P + V + VE                                    & 95.8\%              & 91.0\%                   & 83.5\%                      & 77.8\%             & 70.3\% \\
    P + V + IE + VE                               & 96.8\%              & 92.6\%                   & 83.0\%                      & 82.9\%             & 75.2\% \\
    P + V + IE + VE + $\max$                      & 97.0\%              & \underline{92.8\%}       & 82.1\%                      & \textbf{85.1\%}    & \textbf{76.7\%} \\
    \bottomrule
  \end{tabular}}
  \end{center}
\end{table}

\paragraph{Logistic Regression}
When incorporating the logistic regression model, we conducted experiments to assess the impact of varying amounts of additional training data extracted from the testing set. Specifically, we explored using 1-5\%, 10\%, 20\%, 50\%, 90\% of the frames for the training. We used both the configuration using only the four basic features and the configuration also using the additional $\max$ feature. The results, as presented in \cref{tab:logistic_ablation}, indicate that the  amount of extra training data has minimal effects, as long as there is some extra data. We repeated each configuration 100 times and reported both the mean and standard deviation values. The consistently low standard deviation values observed across all configurations and datasets underscore the robustness of our approach.
We chose $2\%$ extra data with the $\max$ feature as the optimal trade-off between extra data and efficacy.

\begin{table}[ht]
  \caption{Performance comparison between various model configurations, with different amounts of training data for the logistic regression model. $\alpha$ represents the proportion of test set frames employed for the training, with these frames excluded from model evaluation. We repeat the process 100 times, and both mean and standard deviation values are reported. The first half uses the basic four features, and the second half also uses the extra $\max$ feature. Best and second best results highlighted in bold and underlined, respectively. The minimal difference in results between different values of $\alpha>0\%$ is evident.}
  \label{tab:logistic_ablation}
  \begin{center}
  {\small \begin{tabular}{@{}lccccc@{}}
    \toprule
    Configuration          & Ped2                       & Avenue                     & ShanghaiTech               & \ADset                     & \VIOset                    \\
    \midrule
    $\alpha=0\%$           & 96.8\%                     & 92.6\%                     & 83.0\%                     & 82.9\%                     & 75.2\%                     \\
    $\alpha=1\%$           & 98.5\% ± 1.1\%             & 92.5\% ± 0.6\%             & 84.5\% ± 0.6\%             & 89.6\% ± 0.6\%             & 79.6\% ± 0.3\%             \\
    $\alpha=2\%$           & 99.0\% ± 0.6\%             & 92.7\% ± 0.7\%             & 84.7\% ± 0.4\%             & 89.9\% ± 0.4\%             & 79.7\% ± 0.2\%             \\
    $\alpha=3\%$           & 99.2\% ± 0.5\%             & 92.7\% ± 0.7\%             & 84.8\% ± 0.4\%             & 89.9\% ± 0.3\%             & 79.7\% ± 0.1\%             \\
    $\alpha=4\%$           & 99.4\% ± 0.3\%             & 92.7\% ± 0.7\%             & 84.7\% ± 0.3\%             & 89.9\% ± 0.3\%             & 79.8\% ± 0.1\%             \\
    $\alpha=5\%$           & 99.4\% ± 0.4\%             & 93.0\% ± 0.6\%             & 84.8\% ± 0.3\%             & 90.0\% ± 0.2\%             & 79.8\% ± 0.1\%             \\
    $\alpha=10\%$          & 99.5\% ± 0.2\%             & 92.9\% ± 0.6\%             & 84.8\% ± 0.2\%             & 90.1\% ± 0.2\%             & 79.8\% ± 0.1\%             \\
    $\alpha=20\%$          & 99.6\% ± 0.1\%             & 93.1\% ± 0.5\%             & 84.8\% ± 0.2\%             & 90.2\% ± 0.2\%             & 79.8\% ± 0.1\%             \\
    $\alpha=50\%$          & \textbf{99.7\% ± 0.1\%}    & \underline{93.1\% ± 0.3\%} & 84.8\% ± 0.2\%             & 90.2\% ± 0.2\%             & 79.7\% ± 0.3\%             \\
    $\alpha=90\%$          & 99.7\% ± 0.3\%             & \textbf{93.2\% ± 0.6\%}    & 84.8\% ± 0.7\%             & 90.2\% ± 0.5\%             & 79.8\% ± 0.8\%             \\
    \midrule
    $\alpha=0\%$  + $\max$ & 97.0\%                     & 92.8\%                     & 82.1\%                     & 85.1\%                     & 76.7\% \\
    $\alpha=1\%$  + $\max$ & 98.5\% ± 0.8\%             & 92.5\% ± 0.6\%             & 84.5\% ± 0.6\%             & 89.8\% ± 0.5\%             & \underline{80.2\% ± 0.3\%} \\
    $\alpha=2\%$  + $\max$ & 99.0\% ± 0.5\%             & 92.9\% ± 0.5\%             & 84.8\% ± 0.4\%             & 90.0\% ± 0.4\%             & \textbf{80.2\% ± 0.2\%}    \\
    $\alpha=3\%$  + $\max$ & 99.1\% ± 0.7\%             & 92.9\% ± 0.6\%             & 85.0\% ± 0.3\%             & 90.0\% ± 0.3\%             & \textbf{80.2\% ± 0.2\%}    \\
    $\alpha=4\%$  + $\max$ & 99.3\% ± 0.5\%             & 93.0\% ± 0.5\%             & 85.1\% ± 0.3\%             & 90.1\% ± 0.3\%             & \textbf{80.2\% ± 0.2\%}    \\
    $\alpha=5\%$  + $\max$ & 99.3\% ± 0.4\%             & 93.0\% ± 0.4\%             & 85.1\% ± 0.3\%             & 90.2\% ± 0.3\%             & \textbf{80.2\% ± 0.2\%}    \\
    $\alpha=10\%$ + $\max$ & 99.5\% ± 0.2\%             & 93.0\% ± 0.4\%             & 85.2\% ± 0.2\%             & 90.2\% ± 0.2\%             & \textbf{80.2\% ± 0.2\%}    \\
    $\alpha=20\%$ + $\max$ & 99.6\% ± 0.1\%             & 93.0\% ± 0.3\%             & 85.2\% ± 0.2\%             & 90.3\% ± 0.2\%             & 80.1\% ± 0.2\%             \\
    $\alpha=50\%$ + $\max$ & \textbf{99.7\% ± 0.1\%}    & 93.0\% ± 0.3\%             & \textbf{85.3\% ± 0.2\%}    & \textbf{90.4\% ± 0.2\%}    & 80.1\% ± 0.3\%             \\
    $\alpha=90\%$ + $\max$ & \underline{99.7\% ± 0.2\%} & 93.1\% ± 0.6\%             & \underline{85.3\% ± 0.6\%} & \underline{90.4\% ± 0.5\%} & 80.1\% ± 0.8\%             \\
    \bottomrule
  \end{tabular}}
  \end{center}
\end{table}

\section{Conclusion}
\label{sec:conclusion}

Our paper introduces a broader interpretation of anomalies, encompassing both simple anomalies, commonly found in existing benchmarks, and multi-frame complex anomalies. Building upon the foundation laid by AI-VAD \citep{reiss_attribute-based_2022}, we present a novel method that achieves state-of-the-art performance on our proposed datasets while remaining competitive with recent methods on benchmark datasets. We introduce two new datasets of varying complexity, designed to assess the ability of future models to detect complex action-based anomalies.

In future work, we aim to explore even more intricate types of anomalies, such as location and time-based anomalies (e.g. detecting normal actions occurring at abnormal locations or times) thus further advancing the field of general anomaly detection in videos.



\bibliography{references}

\begin{thebibliography}{47}
\providecommand{\natexlab}[1]{#1}
\providecommand{\url}[1]{\texttt{#1}}
\expandafter\ifx\csname urlstyle\endcsname\relax
  \providecommand{\doi}[1]{doi: #1}\else
  \providecommand{\doi}{doi: \begingroup \urlstyle{rm}\Url}\fi

\bibitem[Acsintoae et~al.(2022)Acsintoae, Florescu, Georgescu, Mare, Sumedrea, Ionescu, Khan, and Shah]{acsintoae_ubnormal_2022}
Andra Acsintoae, Andrei Florescu, Mariana-Iuliana Georgescu, Tudor Mare, Paul Sumedrea, Radu~Tudor Ionescu, Fahad~Shahbaz Khan, and Mubarak Shah.
\newblock {UBnormal}: {New} {Benchmark} for {Supervised} {Open}-{Set} {Video} {Anomaly} {Detection}.
\newblock In \emph{Proceedings of the {IEEE}/{CVF} {Conference} on {Computer} {Vision} and {Pattern} {Recognition}}, pp.\  20143--20153, 2022.
\newblock URL \url{https://openaccess.thecvf.com/content/CVPR2022/html/Acsintoae_UBnormal_New_Benchmark_for_Supervised_Open-Set_Video_Anomaly_Detection_CVPR_2022_paper.html}.

\bibitem[Adam et~al.(2008)Adam, Rivlin, Shimshoni, and Reinitz]{adam_robust_2008}
Amit Adam, Ehud Rivlin, Ilan Shimshoni, and Daviv Reinitz.
\newblock Robust real-time unusual event detection using multiple fixed-location monitors.
\newblock \emph{IEEE transactions on pattern analysis and machine intelligence}, 30\penalty0 (3):\penalty0 555--560, March 2008.
\newblock ISSN 0162-8828.
\newblock \doi{10.1109/TPAMI.2007.70825}.

\bibitem[Barbalau et~al.(2023)Barbalau, Ionescu, Georgescu, Dueholm, Ramachandra, Nasrollahi, Khan, Moeslund, and Shah]{barbalau_ssmtl_2023}
Antonio Barbalau, Radu~Tudor Ionescu, Mariana-Iuliana Georgescu, Jacob Dueholm, Bharathkumar Ramachandra, Kamal Nasrollahi, Fahad~Shahbaz Khan, Thomas~B. Moeslund, and Mubarak Shah.
\newblock {SSMTL}++: {Revisiting} {Self}-{Supervised} {Multi}-{Task} {Learning} for {Video} {Anomaly} {Detection}, February 2023.
\newblock URL \url{http://arxiv.org/abs/2207.08003}.
\newblock arXiv:2207.08003 [cs].

\bibitem[Caba~Heilbron et~al.(2015)Caba~Heilbron, Escorcia, Ghanem, and Carlos~Niebles]{caba_heilbron_activitynet_2015}
Fabian Caba~Heilbron, Victor Escorcia, Bernard Ghanem, and Juan Carlos~Niebles.
\newblock {ActivityNet}: {A} {Large}-{Scale} {Video} {Benchmark} for {Human} {Activity} {Understanding}.
\newblock In \emph{Proceedings of the {IEEE} {Conference} on {Computer} {Vision} and {Pattern} {Recognition}}, pp.\  961--970, 2015.
\newblock URL \url{https://openaccess.thecvf.com/content_cvpr_2015/html/Heilbron_ActivityNet_A_Large-Scale_2015_CVPR_paper.html}.

\bibitem[Cai et~al.(2021)Cai, Zhang, Liu, Gao, and Hao]{cai_appearance-motion_2021}
Ruichu Cai, Hao Zhang, Wen Liu, Shenghua Gao, and Zhifeng Hao.
\newblock Appearance-{Motion} {Memory} {Consistency} {Network} for {Video} {Anomaly} {Detection}.
\newblock \emph{Proceedings of the AAAI Conference on Artificial Intelligence}, 35\penalty0 (2):\penalty0 938--946, May 2021.
\newblock ISSN 2374-3468.
\newblock \doi{10.1609/aaai.v35i2.16177}.
\newblock URL \url{https://ojs.aaai.org/index.php/AAAI/article/view/16177}.
\newblock Number: 2.

\bibitem[Cao et~al.(2023)Cao, Lu, Wang, and Zhang]{cao_new_2023}
Congqi Cao, Yue Lu, Peng Wang, and Yanning Zhang.
\newblock A {New} {Comprehensive} {Benchmark} for {Semi}-{Supervised} {Video} {Anomaly} {Detection} and {Anticipation}.
\newblock In \emph{Proceedings of the {IEEE}/{CVF} {Conference} on {Computer} {Vision} and {Pattern} {Recognition}}, pp.\  20392--20401, 2023.
\newblock URL \url{https://openaccess.thecvf.com/content/CVPR2023/html/Cao_A_New_Comprehensive_Benchmark_for_Semi-Supervised_Video_Anomaly_Detection_and_CVPR_2023_paper}.

\bibitem[Carreira et~al.(2018)Carreira, Noland, Banki-Horvath, Hillier, and Zisserman]{carreira_short_2018}
Joao Carreira, Eric Noland, Andras Banki-Horvath, Chloe Hillier, and Andrew Zisserman.
\newblock A {Short} {Note} about {Kinetics}-600, August 2018.
\newblock URL \url{http://arxiv.org/abs/1808.01340}.
\newblock arXiv:1808.01340 [cs].

\bibitem[Chaudhry et~al.(2009)Chaudhry, Ravichandran, Hager, and Vidal]{chaudhry_histograms_2009}
Rizwan Chaudhry, Avinash Ravichandran, Gregory Hager, and Rene Vidal.
\newblock Histograms of oriented optical flow and {Binet}-{Cauchy} kernels on nonlinear dynamical systems for the recognition of human actions.
\newblock In \emph{2009 {IEEE} {Conference} on {Computer} {Vision} and {Pattern} {Recognition}}, pp.\  1932--1939, June 2009.
\newblock \doi{10.1109/CVPR.2009.5206821}.
\newblock ISSN: 1063-6919.

\bibitem[Colque et~al.(2017)Colque, Caetano, de~Andrade, and Schwartz]{colque_histograms_2017}
Rensso Victor Hugo~Mora Colque, Carlos Caetano, Matheus Toledo~Lustosa de~Andrade, and William~Robson Schwartz.
\newblock Histograms of {Optical} {Flow} {Orientation} and {Magnitude} and {Entropy} to {Detect} {Anomalous} {Events} in {Videos}.
\newblock \emph{IEEE Transactions on Circuits and Systems for Video Technology}, 27\penalty0 (3):\penalty0 673--682, March 2017.
\newblock ISSN 1558-2205.
\newblock \doi{10.1109/TCSVT.2016.2637778}.
\newblock Conference Name: IEEE Transactions on Circuits and Systems for Video Technology.

\bibitem[Eskin et~al.(2002)Eskin, Arnold, Prerau, Portnoy, and Stolfo]{eskin_geometric_2002}
Eleazar Eskin, Andrew Arnold, Michael Prerau, Leonid Portnoy, and Sal Stolfo.
\newblock A {Geometric} {Framework} for {Unsupervised} {Anomaly} {Detection}.
\newblock In Daniel Barbará and Sushil Jajodia (eds.), \emph{Applications of {Data} {Mining} in {Computer} {Security}}, Advances in {Information} {Security}, pp.\  77--101. Springer US, Boston, MA, 2002.
\newblock ISBN 978-1-4615-0953-0.
\newblock \doi{10.1007/978-1-4615-0953-0_4}.
\newblock URL \url{https://doi.org/10.1007/978-1-4615-0953-0_4}.

\bibitem[Fan et~al.(2018)Fan, Wen, Li, Qiu, and Levine]{fan_video_2018}
Yaxiang Fan, Gongjian Wen, Deren Li, Shaohua Qiu, and Martin~D. Levine.
\newblock Video {Anomaly} {Detection} and {Localization} via {Gaussian} {Mixture} {Fully} {Convolutional} {Variational} {Autoencoder}, May 2018.
\newblock URL \url{http://arxiv.org/abs/1805.11223}.
\newblock arXiv:1805.11223 [cs].

\bibitem[Georgescu et~al.(2021)Georgescu, Barbalau, Ionescu, Khan, Popescu, and Shah]{georgescu_anomaly_2021}
Mariana-Iuliana Georgescu, Antonio Barbalau, Radu~Tudor Ionescu, Fahad~Shahbaz Khan, Marius Popescu, and Mubarak Shah.
\newblock Anomaly {Detection} in {Video} via {Self}-{Supervised} and {Multi}-{Task} {Learning}.
\newblock In \emph{Proceedings of the {IEEE}/{CVF} {Conference} on {Computer} {Vision} and {Pattern} {Recognition}}, pp.\  12742--12752, 2021.
\newblock URL \url{https://openaccess.thecvf.com/content/CVPR2021/html/Georgescu_Anomaly_Detection_in_Video_via_Self-Supervised_and_Multi-Task_Learning_CVPR_2021_paper.html}.

\bibitem[Georgescu et~al.(2022)Georgescu, Ionescu, Khan, Popescu, and Shah]{georgescu_background-agnostic_2022}
Mariana~Iuliana Georgescu, Radu~Tudor Ionescu, Fahad~Shahbaz Khan, Marius Popescu, and Mubarak Shah.
\newblock A {Background}-{Agnostic} {Framework} {With} {Adversarial} {Training} for {Abnormal} {Event} {Detection} in {Video}.
\newblock \emph{IEEE Transactions on Pattern Analysis and Machine Intelligence}, 44\penalty0 (9):\penalty0 4505--4523, September 2022.
\newblock ISSN 1939-3539.
\newblock \doi{10.1109/TPAMI.2021.3074805}.
\newblock Conference Name: IEEE Transactions on Pattern Analysis and Machine Intelligence.

\bibitem[Gidaris et~al.(2018)Gidaris, Singh, and Komodakis]{gidaris_unsupervised_2018}
Spyros Gidaris, Praveer Singh, and Nikos Komodakis.
\newblock Unsupervised {Representation} {Learning} by {Predicting} {Image} {Rotations}, March 2018.
\newblock URL \url{http://arxiv.org/abs/1803.07728}.
\newblock arXiv:1803.07728 [cs].

\bibitem[Glodek et~al.(2013)Glodek, Schels, and Schwenker]{glodek_ensemble_2013}
Michael Glodek, Martin Schels, and Friedhelm Schwenker.
\newblock Ensemble {Gaussian} mixture models for probability density estimation.
\newblock \emph{Computational Statistics}, 27:\penalty0 127--138, December 2013.
\newblock \doi{10.1007/s00180-012-0374-5}.

\bibitem[Gong et~al.(2019)Gong, Liu, Le, Saha, Mansour, Venkatesh, and Hengel]{gong_memorizing_2019}
Dong Gong, Lingqiao Liu, Vuong Le, Budhaditya Saha, Moussa~Reda Mansour, Svetha Venkatesh, and Anton van~den Hengel.
\newblock Memorizing {Normality} to {Detect} {Anomaly}: {Memory}-augmented {Deep} {Autoencoder} for {Unsupervised} {Anomaly} {Detection}, August 2019.
\newblock URL \url{http://arxiv.org/abs/1904.02639}.
\newblock arXiv:1904.02639 [cs].

\bibitem[Goyal et~al.(2017)Goyal, Ebrahimi~Kahou, Michalski, Materzynska, Westphal, Kim, Haenel, Fruend, Yianilos, Mueller-Freitag, Hoppe, Thurau, Bax, and Memisevic]{goyal_something_2017}
Raghav Goyal, Samira Ebrahimi~Kahou, Vincent Michalski, Joanna Materzynska, Susanne Westphal, Heuna Kim, Valentin Haenel, Ingo Fruend, Peter Yianilos, Moritz Mueller-Freitag, Florian Hoppe, Christian Thurau, Ingo Bax, and Roland Memisevic.
\newblock The "{Something} {Something}" {Video} {Database} for {Learning} and {Evaluating} {Visual} {Common} {Sense}.
\newblock In \emph{Proceedings of the {IEEE} {International} {Conference} on {Computer} {Vision}}, pp.\  5842--5850, 2017.
\newblock URL \url{https://openaccess.thecvf.com/content_iccv_2017/html/Goyal_The_Something_Something_ICCV_2017_paper.html}.

\bibitem[Hasan et~al.(2016)Hasan, Choi, Neumann, Roy-Chowdhury, and Davis]{hasan_learning_2016}
Mahmudul Hasan, Jonghyun Choi, Jan Neumann, Amit~K. Roy-Chowdhury, and Larry~S. Davis.
\newblock Learning {Temporal} {Regularity} in {Video} {Sequences}.
\newblock In \emph{Proceedings of the {IEEE} {Conference} on {Computer} {Vision} and {Pattern} {Recognition}}, pp.\  733--742, 2016.
\newblock URL \url{https://openaccess.thecvf.com/content_cvpr_2016/html/Hasan_Learning_Temporal_Regularity_CVPR_2016_paper}.

\bibitem[Kay et~al.(2017)Kay, Carreira, Simonyan, Zhang, Hillier, Vijayanarasimhan, Viola, Green, Back, Natsev, Suleyman, and Zisserman]{kay_kinetics_2017}
Will Kay, Joao Carreira, Karen Simonyan, Brian Zhang, Chloe Hillier, Sudheendra Vijayanarasimhan, Fabio Viola, Tim Green, Trevor Back, Paul Natsev, Mustafa Suleyman, and Andrew Zisserman.
\newblock The {Kinetics} {Human} {Action} {Video} {Dataset}, May 2017.
\newblock URL \url{http://arxiv.org/abs/1705.06950}.
\newblock arXiv:1705.06950 [cs].

\bibitem[Kuehne et~al.(2011)Kuehne, Jhuang, Garrote, Poggio, and Serre]{kuehne_hmdb_2011}
H~Kuehne, H~Jhuang, E~Garrote, T~Poggio, and T~Serre.
\newblock {HMDB}: {A} {Large} {Video} {Database} for {Human} {Motion} {Recognition}.
\newblock In \emph{Proceedings of the {International} {Conference} on {Computer} {Vision} ({ICCV})}, 2011.

\bibitem[Latecki et~al.(2007)Latecki, Lazarevic, and Pokrajac]{latecki_outlier_2007}
Longin~Jan Latecki, Aleksandar Lazarevic, and Dragoljub Pokrajac.
\newblock Outlier {Detection} with {Kernel} {Density} {Functions}.
\newblock In Petra Perner (ed.), \emph{Machine {Learning} and {Data} {Mining} in {Pattern} {Recognition}}, Lecture {Notes} in {Computer} {Science}, pp.\  61--75, Berlin, Heidelberg, 2007. Springer.
\newblock ISBN 978-3-540-73499-4.
\newblock \doi{10.1007/978-3-540-73499-4_6}.

\bibitem[Liu et~al.(2018)Liu, Luo, Lian, and Gao]{liu_future_2018}
Wen Liu, Weixin Luo, Dongze Lian, and Shenghua Gao.
\newblock Future {Frame} {Prediction} for {Anomaly} {Detection} – {A} {New} {Baseline}.
\newblock In \emph{Proceedings of the {IEEE} {Conference} on {Computer} {Vision} and {Pattern} {Recognition}}, pp.\  6536--6545, 2018.
\newblock URL \url{https://openaccess.thecvf.com/content_cvpr_2018/html/Liu_Future_Frame_Prediction_CVPR_2018_paper}.

\bibitem[Liu et~al.(2021)Liu, Nie, Long, Zhang, and Li]{liu_hybrid_2021}
Zhian Liu, Yongwei Nie, Chengjiang Long, Qing Zhang, and Guiqing Li.
\newblock A {Hybrid} {Video} {Anomaly} {Detection} {Framework} via {Memory}-{Augmented} {Flow} {Reconstruction} and {Flow}-{Guided} {Frame} {Prediction}.
\newblock In \emph{Proceedings of the {IEEE}/{CVF} {International} {Conference} on {Computer} {Vision}}, pp.\  13588--13597, 2021.
\newblock URL \url{https://openaccess.thecvf.com/content/ICCV2021/html/Liu_A_Hybrid_Video_Anomaly_Detection_Framework_via_Memory-Augmented_Flow_Reconstruction_ICCV_2021_paper}.

\bibitem[Lowe(2004)]{lowe_distinctive_2004}
David~G. Lowe.
\newblock Distinctive {Image} {Features} from {Scale}-{Invariant} {Keypoints}.
\newblock \emph{International Journal of Computer Vision}, 60\penalty0 (2):\penalty0 91--110, November 2004.
\newblock ISSN 1573-1405.
\newblock \doi{10.1023/B:VISI.0000029664.99615.94}.
\newblock URL \url{https://doi.org/10.1023/B:VISI.0000029664.99615.94}.

\bibitem[Lu et~al.(2013)Lu, Shi, and Jia]{lu_abnormal_2013}
Cewu Lu, Jianping Shi, and Jiaya Jia.
\newblock Abnormal {Event} {Detection} at 150 {FPS} in {MATLAB}.
\newblock In \emph{2013 {IEEE} {International} {Conference} on {Computer} {Vision}}, pp.\  2720--2727, December 2013.
\newblock \doi{10.1109/ICCV.2013.338}.
\newblock ISSN: 2380-7504.

\bibitem[Lu et~al.(2019)Lu, Kumar, Nabavi, and Wang]{lu_future_2019}
Yiwei Lu, K~Mahesh Kumar, Seyed~shahabeddin Nabavi, and Yang Wang.
\newblock Future {Frame} {Prediction} {Using} {Convolutional} {VRNN} for {Anomaly} {Detection}.
\newblock In \emph{2019 16th {IEEE} {International} {Conference} on {Advanced} {Video} and {Signal} {Based} {Surveillance} ({AVSS})}, pp.\  1--8, September 2019.
\newblock \doi{10.1109/AVSS.2019.8909850}.
\newblock ISSN: 2643-6213.

\bibitem[Luo et~al.(2017)Luo, Liu, and Gao]{luo_remembering_2017}
Weixin Luo, Wen Liu, and Shenghua Gao.
\newblock Remembering history with convolutional {LSTM} for anomaly detection.
\newblock In \emph{2017 {IEEE} {International} {Conference} on {Multimedia} and {Expo} ({ICME})}, pp.\  439--444, July 2017.
\newblock \doi{10.1109/ICME.2017.8019325}.
\newblock ISSN: 1945-788X.

\bibitem[Lv et~al.(2021)Lv, Chen, Cui, Xu, Li, and Yang]{lv_learning_2021}
Hui Lv, Chen Chen, Zhen Cui, Chunyan Xu, Yong Li, and Jian Yang.
\newblock Learning {Normal} {Dynamics} in {Videos} with {Meta} {Prototype} {Network}, May 2021.
\newblock URL \url{http://arxiv.org/abs/2104.06689}.
\newblock arXiv:2104.06689 [cs].

\bibitem[Mahadevan et~al.(2010)Mahadevan, Li, Bhalodia, and Vasconcelos]{mahadevan_anomaly_2010}
Vijay Mahadevan, Weixin Li, Viral Bhalodia, and Nuno Vasconcelos.
\newblock Anomaly detection in crowded scenes.
\newblock In \emph{2010 {IEEE} {Computer} {Society} {Conference} on {Computer} {Vision} and {Pattern} {Recognition}}, pp.\  1975--1981, San Francisco, CA, USA, June 2010. IEEE.
\newblock ISBN 978-1-4244-6984-0.
\newblock \doi{10.1109/CVPR.2010.5539872}.
\newblock URL \url{http://ieeexplore.ieee.org/document/5539872/}.

\bibitem[Nguyen \& Meunier(2019)Nguyen and Meunier]{nguyen_anomaly_2019}
Trong-Nguyen Nguyen and Jean Meunier.
\newblock Anomaly {Detection} in {Video} {Sequence} {With} {Appearance}-{Motion} {Correspondence}.
\newblock In \emph{Proceedings of the {IEEE}/{CVF} {International} {Conference} on {Computer} {Vision}}, pp.\  1273--1283, 2019.
\newblock URL \url{https://openaccess.thecvf.com/content_ICCV_2019/html/Nguyen_Anomaly_Detection_in_Video_Sequence_With_Appearance-Motion_Correspondence_ICCV_2019_paper}.

\bibitem[of~Minnesota(2006)]{university_of_minnesota_unusual_2006}
University of~Minnesota.
\newblock Unusual crowd activity dataset of university of minnesota, 2006.
\newblock URL \url{http://mha.cs.umn.edu/proj_events.shtml}.

\bibitem[Park et~al.(2020)Park, Noh, and Ham]{park_learning_2020}
Hyunjong Park, Jongyoun Noh, and Bumsub Ham.
\newblock Learning {Memory}-{Guided} {Normality} for {Anomaly} {Detection}.
\newblock In \emph{Proceedings of the {IEEE}/{CVF} {Conference} on {Computer} {Vision} and {Pattern} {Recognition}}, pp.\  14372--14381, 2020.
\newblock URL \url{https://openaccess.thecvf.com/content_CVPR_2020/html/Park_Learning_Memory-Guided_Normality_for_Anomaly_Detection_CVPR_2020_paper}.

\bibitem[Perš et~al.(2010)Perš, Sulić, Kristan, Perše, Polanec, and Kovačič]{pers_histograms_2010}
Janez Perš, Vildana Sulić, Matej Kristan, Matej Perše, Klemen Polanec, and Stanislav Kovačič.
\newblock Histograms of optical flow for efficient representation of body motion.
\newblock \emph{Pattern Recognition Letters}, 31\penalty0 (11):\penalty0 1369--1376, August 2010.
\newblock ISSN 0167-8655.
\newblock \doi{10.1016/j.patrec.2010.03.024}.
\newblock URL \url{https://www.sciencedirect.com/science/article/pii/S0167865510001121}.

\bibitem[Qualcomm(2018)]{qualcomm_moving_2018}
Qualcomm.
\newblock Moving {Objects} {Dataset}: {Something}-{Something} v. 2, 2018.
\newblock URL \url{https://developer.qualcomm.com/software/ai-datasets/something-something}.

\bibitem[Ramachandra \& Jones(2020)Ramachandra and Jones]{ramachandra_street_2020}
Bharathkumar Ramachandra and Michael~J. Jones.
\newblock Street {Scene}: {A} new dataset and evaluation protocol for video anomaly detection.
\newblock In \emph{2020 {IEEE} {Winter} {Conference} on {Applications} of {Computer} {Vision} ({WACV})}, pp.\  2558--2567, March 2020.
\newblock \doi{10.1109/WACV45572.2020.9093457}.
\newblock ISSN: 2642-9381.

\bibitem[Reiss \& Hoshen(2022)Reiss and Hoshen]{reiss_attribute-based_2022}
Tal Reiss and Yedid Hoshen.
\newblock Attribute-based {Representations} for {Accurate} and {Interpretable} {Video} {Anomaly} {Detection}, December 2022.
\newblock URL \url{http://arxiv.org/abs/2212.00789}.
\newblock arXiv:2212.00789 [cs].

\bibitem[Rodrigues et~al.(2020)Rodrigues, Bhargava, Velmurugan, and Chaudhuri]{rodrigues_multi-timescale_2020}
Royston Rodrigues, Neha Bhargava, Rajbabu Velmurugan, and Subhasis Chaudhuri.
\newblock Multi-timescale {Trajectory} {Prediction} for {Abnormal} {Human} {Activity} {Detection}.
\newblock In \emph{2020 {IEEE} {Winter} {Conference} on {Applications} of {Computer} {Vision} ({WACV})}, pp.\  2615--2623, March 2020.
\newblock \doi{10.1109/WACV45572.2020.9093633}.
\newblock ISSN: 2642-9381.

\bibitem[Sabzalian et~al.(2019)Sabzalian, Marvi, and Ahmadyfard]{sabzalian_deep_2019}
Behnam Sabzalian, Hossein Marvi, and Alireza Ahmadyfard.
\newblock Deep and {Sparse} features {For} {Anomaly} {Detection} and {Localization} in video.
\newblock In \emph{2019 4th {International} {Conference} on {Pattern} {Recognition} and {Image} {Analysis} ({IPRIA})}, pp.\  173--178, March 2019.
\newblock \doi{10.1109/PRIA.2019.8786007}.
\newblock ISSN: 2049-3630.

\bibitem[Smaira et~al.(2020)Smaira, Carreira, Noland, Clancy, Wu, and Zisserman]{smaira_short_2020}
Lucas Smaira, João Carreira, Eric Noland, Ellen Clancy, Amy Wu, and Andrew Zisserman.
\newblock A {Short} {Note} on the {Kinetics}-700-2020 {Human} {Action} {Dataset}, October 2020.
\newblock URL \url{http://arxiv.org/abs/2010.10864}.
\newblock arXiv:2010.10864 [cs].

\bibitem[Soomro et~al.(2012)Soomro, Zamir, and Shah]{soomro_ucf101_2012}
Khurram Soomro, Amir~Roshan Zamir, and Mubarak Shah.
\newblock {UCF101}: {A} {Dataset} of 101 {Human} {Actions} {Classes} {From} {Videos} in {The} {Wild}, December 2012.
\newblock URL \url{http://arxiv.org/abs/1212.0402}.
\newblock arXiv:1212.0402 [cs].

\bibitem[Sun \& Gong(2023)Sun and Gong]{sun_hierarchical_2023}
Shengyang Sun and Xiaojin Gong.
\newblock Hierarchical {Semantic} {Contrast} for {Scene}-{Aware} {Video} {Anomaly} {Detection}.
\newblock In \emph{Proceedings of the {IEEE}/{CVF} {Conference} on {Computer} {Vision} and {Pattern} {Recognition}}, pp.\  22846--22856, 2023.
\newblock URL \url{https://openaccess.thecvf.com/content/CVPR2023/html/Sun_Hierarchical_Semantic_Contrast_for_Scene-Aware_Video_Anomaly_Detection_CVPR_2023_paper}.

\bibitem[Wang et~al.(2022)Wang, Wang, Qin, Zhang, Bao, and Huang]{wang_video_2022}
Guodong Wang, Yunhong Wang, Jie Qin, Dongming Zhang, Xiuguo Bao, and Di~Huang.
\newblock Video {Anomaly} {Detection} by {Solving} {Decoupled} {Spatio}-{Temporal} {Jigsaw} {Puzzles}.
\newblock In Shai Avidan, Gabriel Brostow, Moustapha Cissé, Giovanni~Maria Farinella, and Tal Hassner (eds.), \emph{Computer {Vision} – {ECCV} 2022}, Lecture {Notes} in {Computer} {Science}, pp.\  494--511, Cham, 2022. Springer Nature Switzerland.
\newblock ISBN 978-3-031-20080-9.
\newblock \doi{10.1007/978-3-031-20080-9_29}.

\bibitem[Wang et~al.(2023)Wang, Huang, Zhao, Tong, He, Wang, Wang, and Qiao]{wang_videomae_2023}
Limin Wang, Bingkun Huang, Zhiyu Zhao, Zhan Tong, Yinan He, Yi~Wang, Yali Wang, and Yu~Qiao.
\newblock {VideoMAE} {V2}: {Scaling} {Video} {Masked} {Autoencoders} {With} {Dual} {Masking}.
\newblock In \emph{Proceedings of the {IEEE}/{CVF} {Conference} on {Computer} {Vision} and {Pattern} {Recognition}}, pp.\  14549--14560, 2023.
\newblock URL \url{https://openaccess.thecvf.com/content/CVPR2023/html/Wang_VideoMAE_V2_Scaling_Video_Masked_Autoencoders_With_Dual_Masking_CVPR_2023_paper.html}.

\bibitem[Wei et~al.(2018)Wei, Lim, Zisserman, and Freeman]{wei_learning_2018}
Donglai Wei, Joseph~J. Lim, Andrew Zisserman, and William~T. Freeman.
\newblock Learning and {Using} the {Arrow} of {Time}.
\newblock In \emph{Proceedings of the {IEEE} {Conference} on {Computer} {Vision} and {Pattern} {Recognition}}, pp.\  8052--8060, 2018.
\newblock URL \url{https://openaccess.thecvf.com/content_cvpr_2018/html/Wei_Learning_and_Using_CVPR_2018_paper.html}.

\bibitem[Yang et~al.(2022)Yang, Wu, Liu, and Liu]{yang_dynamic_2022}
Zhiwei Yang, Peng Wu, Jing Liu, and Xiaotao Liu.
\newblock Dynamic {Local} {Aggregation} {Network} with {Adaptive} {Clusterer} for {Anomaly} {Detection}, July 2022.
\newblock URL \url{http://arxiv.org/abs/2207.10948}.
\newblock arXiv:2207.10948 [cs].

\bibitem[Yu et~al.(2020)Yu, Wang, Cai, Zhu, Xu, Yin, and Kloft]{yu_cloze_2020}
Guang Yu, Siqi Wang, Zhiping Cai, En~Zhu, Chuanfu Xu, Jianping Yin, and Marius Kloft.
\newblock Cloze {Test} {Helps}: {Effective} {Video} {Anomaly} {Detection} via {Learning} to {Complete} {Video} {Events}.
\newblock In \emph{Proceedings of the 28th {ACM} {International} {Conference} on {Multimedia}}, {MM} '20, pp.\  583--591, New York, NY, USA, October 2020. Association for Computing Machinery.
\newblock ISBN 978-1-4503-7988-5.
\newblock \doi{10.1145/3394171.3413973}.
\newblock URL \url{https://dl.acm.org/doi/10.1145/3394171.3413973}.

\bibitem[Yu et~al.(2022)Yu, Wang, Cai, Liu, Xu, and Wu]{yu_deep_2022}
Guang Yu, Siqi Wang, Zhiping Cai, Xinwang Liu, Chuanfu Xu, and Chengkun Wu.
\newblock Deep {Anomaly} {Discovery} {From} {Unlabeled} {Videos} via {Normality} {Advantage} and {Self}-{Paced} {Refinement}.
\newblock In \emph{Proceedings of the {IEEE}/{CVF} {Conference} on {Computer} {Vision} and {Pattern} {Recognition}}, pp.\  13987--13998, 2022.
\newblock URL \url{https://openaccess.thecvf.com/content/CVPR2022/html/Yu_Deep_Anomaly_Discovery_From_Unlabeled_Videos_via_Normality_Advantage_and_CVPR_2022_paper.html}.

\end{thebibliography}
\bibliographystyle{iclr2024_conference}

\newpage

\appendix
\section{Samples from proposed datasets}
\subsection{\ADset\ Examples}
\label{app:ad_examples}

\begin{figure}[h]
\begin{center}
  \includegraphics[width=0.1\linewidth]{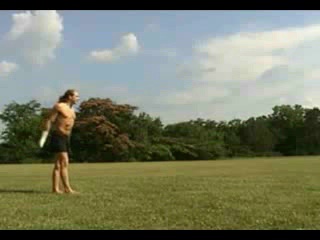}
  \includegraphics[width=0.1\linewidth]{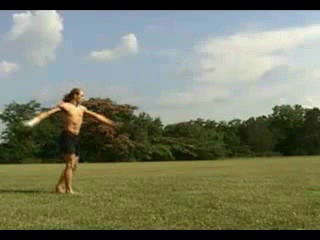}
  \includegraphics[width=0.1\linewidth]{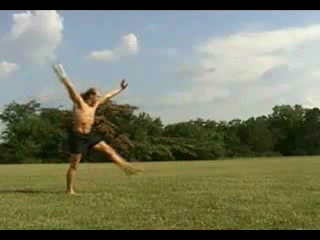}
  \includegraphics[width=0.1\linewidth]{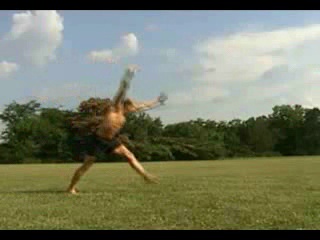}
  \includegraphics[width=0.1\linewidth]{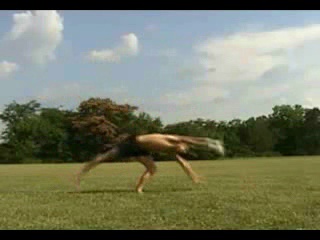}
  \includegraphics[width=0.1\linewidth]{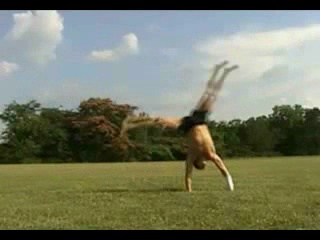}
  \includegraphics[width=0.1\linewidth]{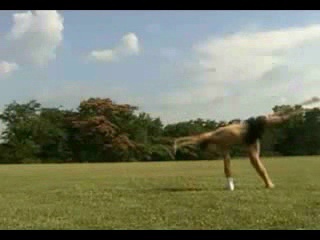}
  \includegraphics[width=0.1\linewidth]{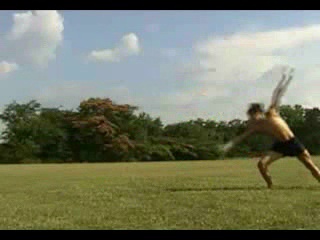}
  \includegraphics[width=0.1\linewidth]{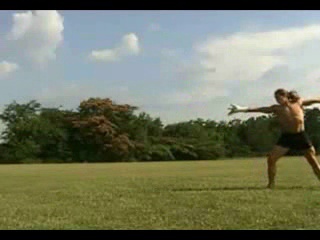} \\
  \includegraphics[width=0.1\linewidth]{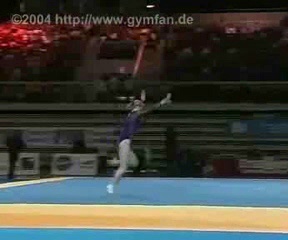}
  \includegraphics[width=0.1\linewidth]{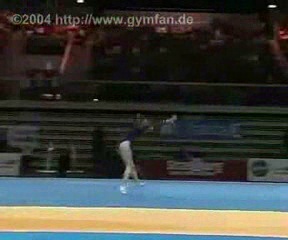}
  \includegraphics[width=0.1\linewidth]{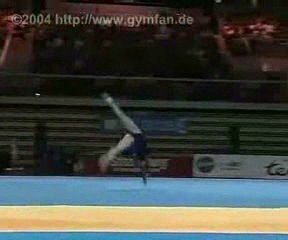}
  \includegraphics[width=0.1\linewidth]{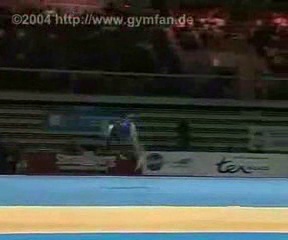}
  \includegraphics[width=0.1\linewidth]{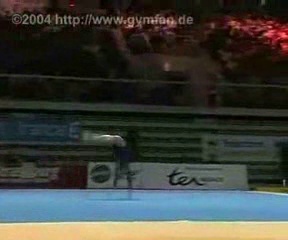} \\
  \includegraphics[width=0.1\linewidth]{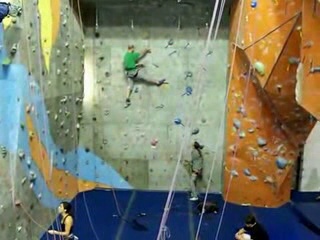}
  \includegraphics[width=0.1\linewidth]{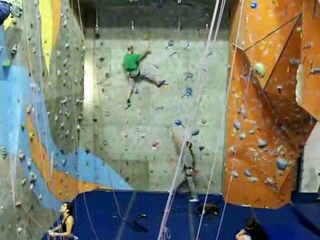}
  \includegraphics[width=0.1\linewidth]{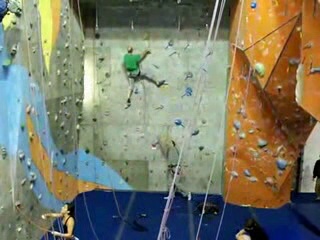}
  \includegraphics[width=0.1\linewidth]{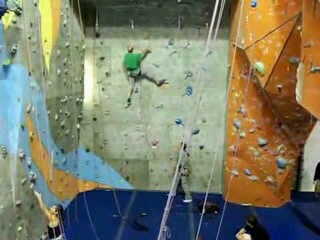}
  \includegraphics[width=0.1\linewidth]{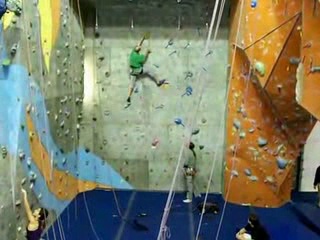}
  \includegraphics[width=0.1\linewidth]{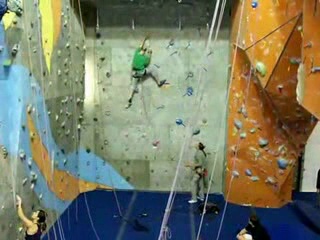}
  \includegraphics[width=0.1\linewidth]{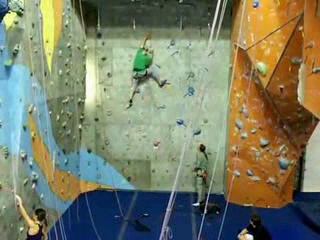}
  \includegraphics[width=0.1\linewidth]{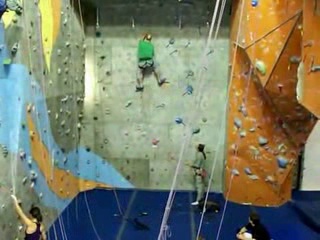}
  \includegraphics[width=0.1\linewidth]{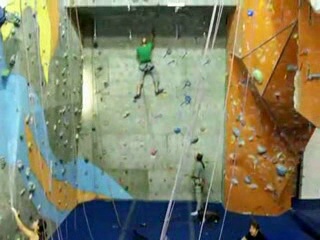} \\
  \includegraphics[width=0.1\linewidth]{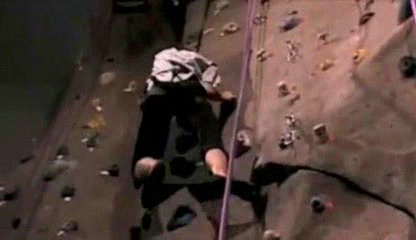}
  \includegraphics[width=0.1\linewidth]{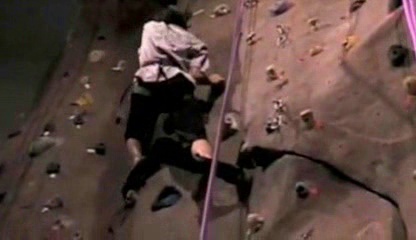}
  \includegraphics[width=0.1\linewidth]{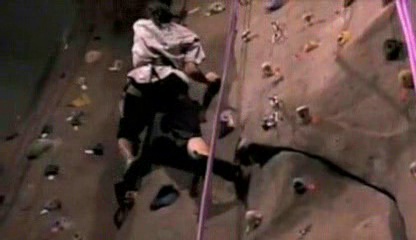}
  \includegraphics[width=0.1\linewidth]{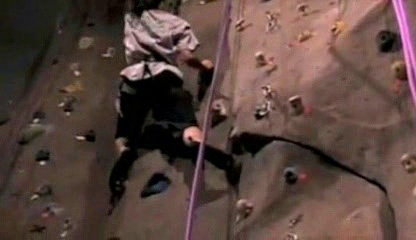}
  \includegraphics[width=0.1\linewidth]{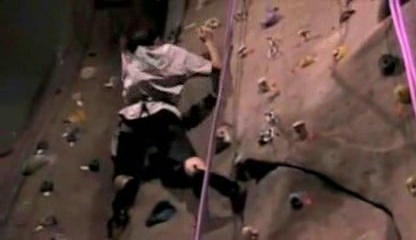}
  \includegraphics[width=0.1\linewidth]{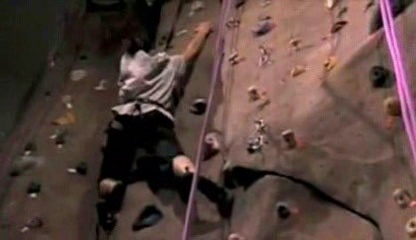}
  \includegraphics[width=0.1\linewidth]{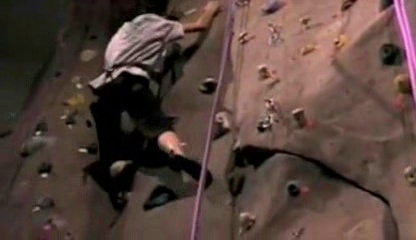}
  \includegraphics[width=0.1\linewidth]{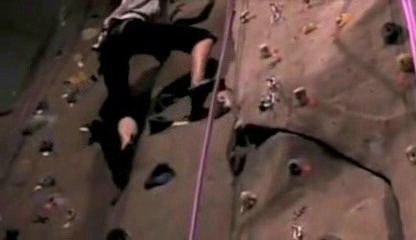}
  \includegraphics[width=0.1\linewidth]{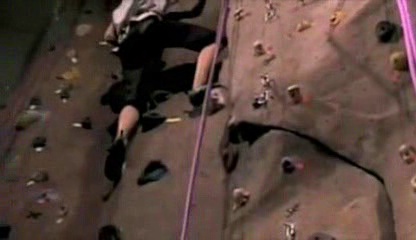} \\
  \includegraphics[width=0.1\linewidth]{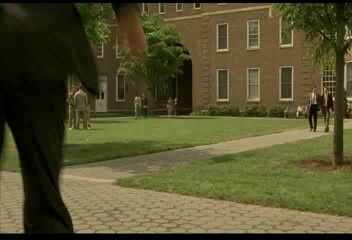}
  \includegraphics[width=0.1\linewidth]{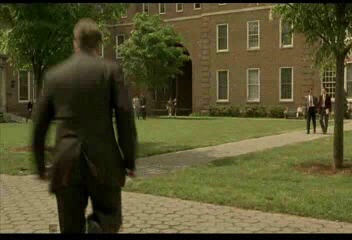}
  \includegraphics[width=0.1\linewidth]{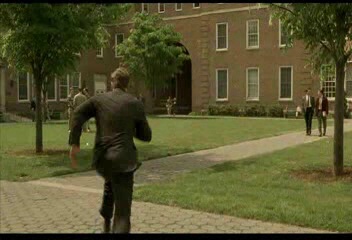}
  \includegraphics[width=0.1\linewidth]{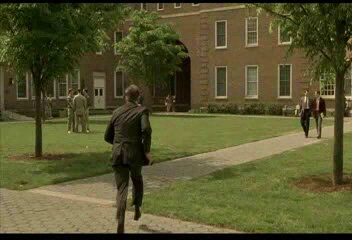}
  \includegraphics[width=0.1\linewidth]{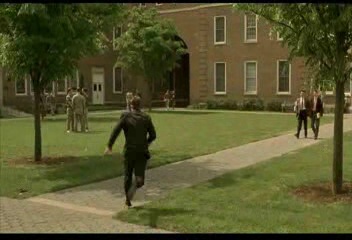}
  \includegraphics[width=0.1\linewidth]{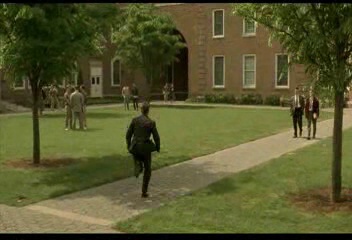}
  \includegraphics[width=0.1\linewidth]{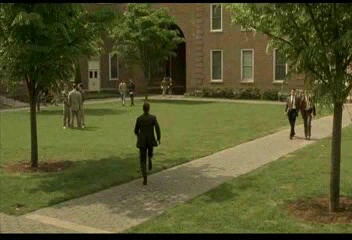}
  \includegraphics[width=0.1\linewidth]{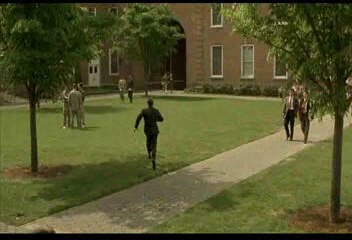}
  \includegraphics[width=0.1\linewidth]{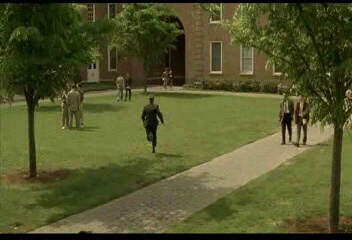} \\
  \includegraphics[width=0.1\linewidth]{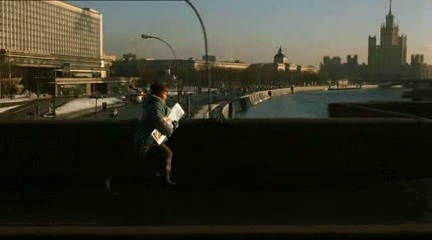}
  \includegraphics[width=0.1\linewidth]{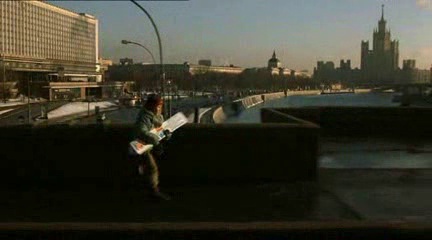}
  \includegraphics[width=0.1\linewidth]{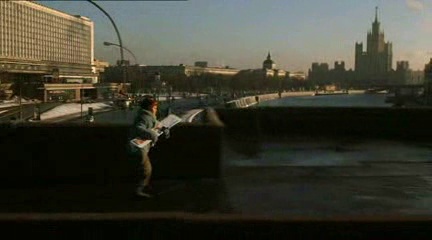}
  \includegraphics[width=0.1\linewidth]{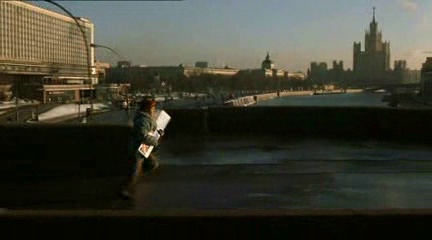}
  \includegraphics[width=0.1\linewidth]{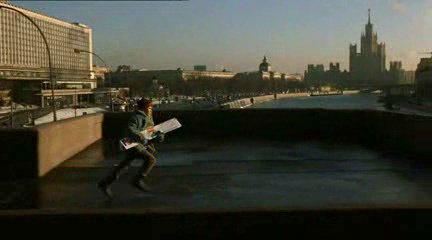}
  \includegraphics[width=0.1\linewidth]{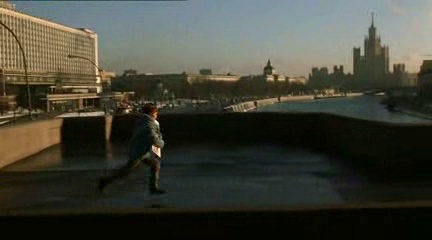}
  \includegraphics[width=0.1\linewidth]{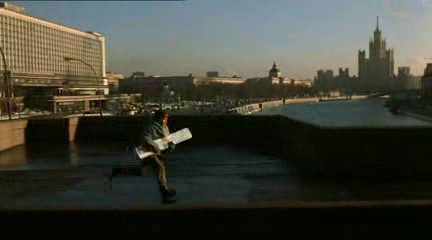}
  \includegraphics[width=0.1\linewidth]{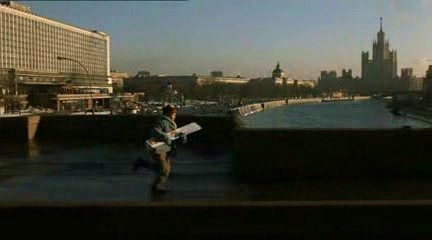}
  \includegraphics[width=0.1\linewidth]{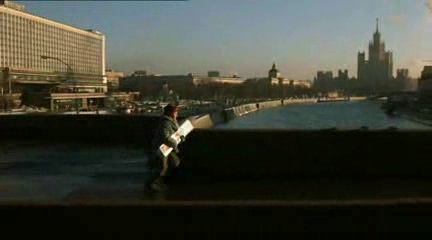} \\
  \includegraphics[width=0.1\linewidth]{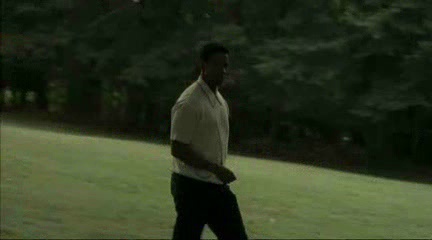}
  \includegraphics[width=0.1\linewidth]{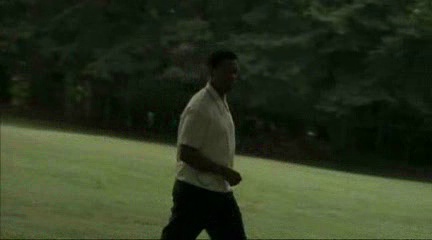}
  \includegraphics[width=0.1\linewidth]{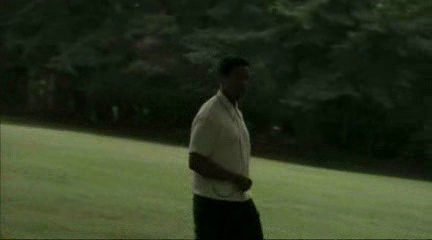}
  \includegraphics[width=0.1\linewidth]{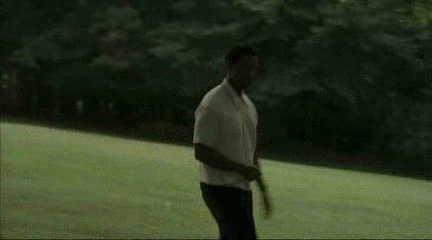}
  \includegraphics[width=0.1\linewidth]{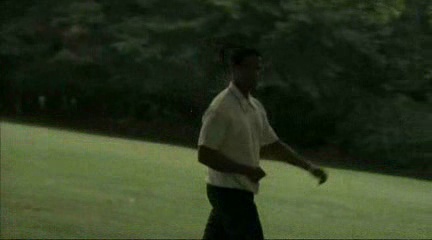}
  \includegraphics[width=0.1\linewidth]{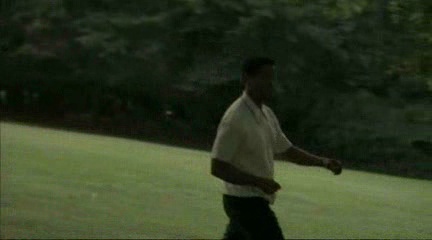}
  \includegraphics[width=0.1\linewidth]{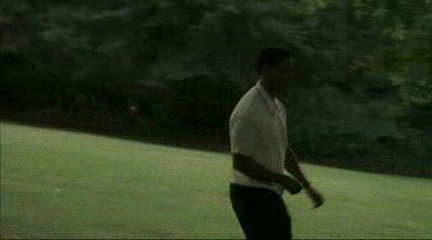}
  \includegraphics[width=0.1\linewidth]{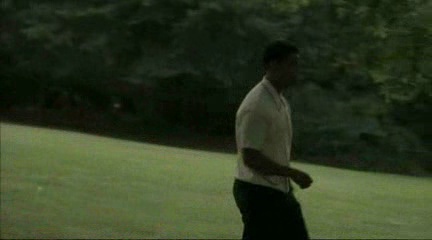} \\
  \includegraphics[width=0.1\linewidth]{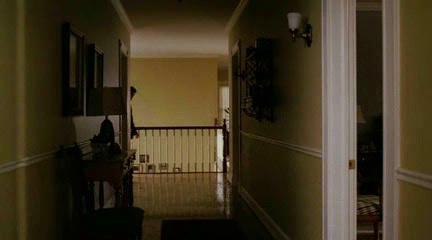}
  \includegraphics[width=0.1\linewidth]{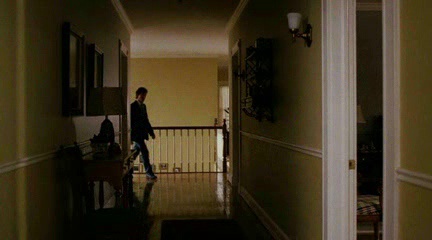}
  \includegraphics[width=0.1\linewidth]{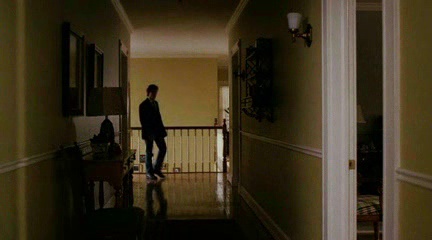}
  \includegraphics[width=0.1\linewidth]{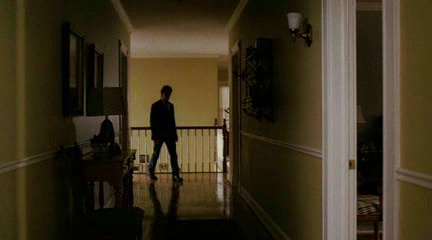}
  \includegraphics[width=0.1\linewidth]{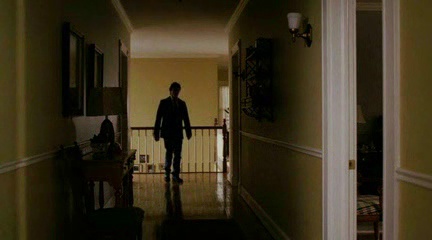}
  \includegraphics[width=0.1\linewidth]{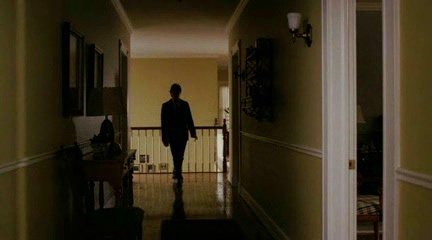}
  \includegraphics[width=0.1\linewidth]{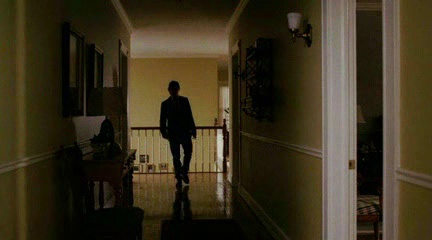}
  \includegraphics[width=0.1\linewidth]{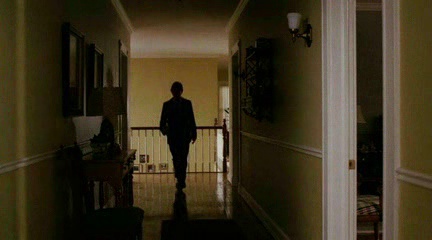}
  \includegraphics[width=0.1\linewidth]{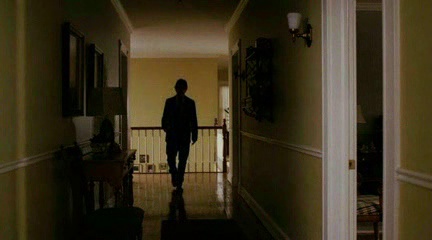}
  \end{center}
  \caption{\ADset\ examples: cartwheel (2), climb (2), run (2), walk (2).}
\end{figure}

\subsection{\VIOset\ Examples}
\label{app:vio_examples}

\begin{figure}[h]
\begin{center}
  \includegraphics[width=0.1\linewidth]{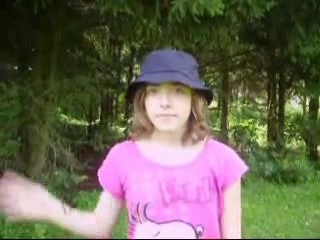}
  \includegraphics[width=0.1\linewidth]{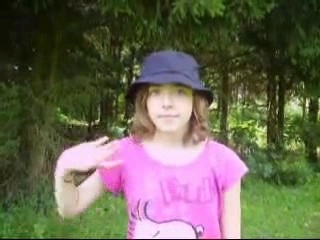}
  \includegraphics[width=0.1\linewidth]{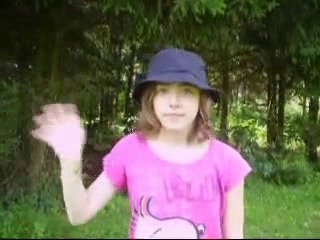}
  \includegraphics[width=0.1\linewidth]{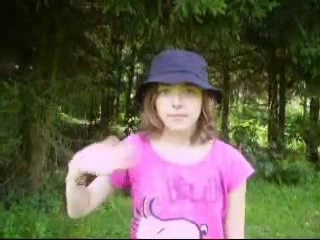}
  \includegraphics[width=0.1\linewidth]{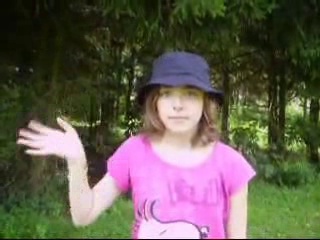}
  \includegraphics[width=0.1\linewidth]{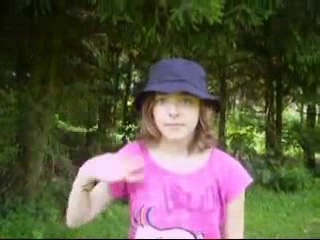}
  \includegraphics[width=0.1\linewidth]{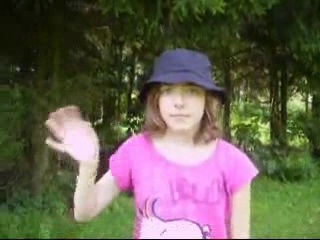}
  \includegraphics[width=0.1\linewidth]{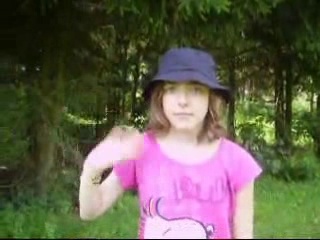}
  \includegraphics[width=0.1\linewidth]{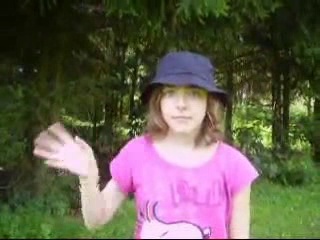} \\
  \includegraphics[width=0.1\linewidth]{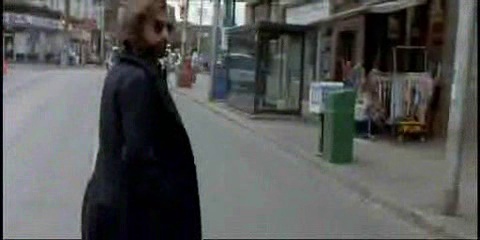}
  \includegraphics[width=0.1\linewidth]{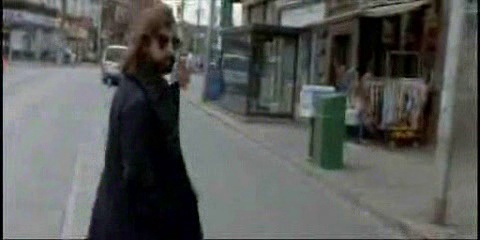}
  \includegraphics[width=0.1\linewidth]{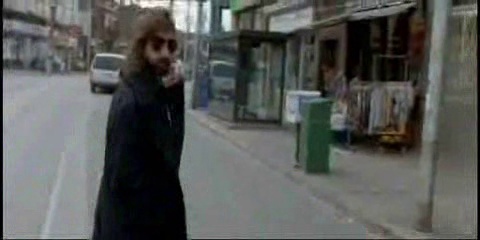}
  \includegraphics[width=0.1\linewidth]{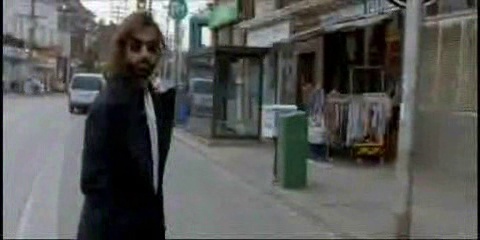}
  \includegraphics[width=0.1\linewidth]{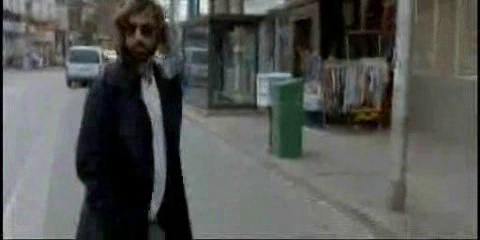}
  \includegraphics[width=0.1\linewidth]{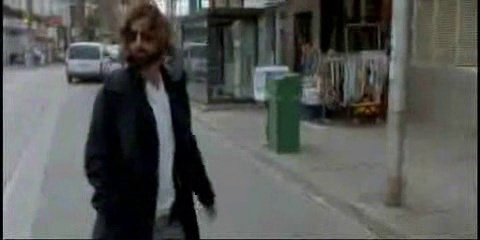}
  \includegraphics[width=0.1\linewidth]{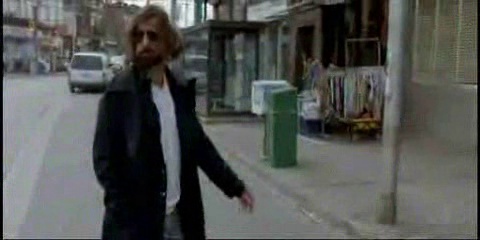}
  \includegraphics[width=0.1\linewidth]{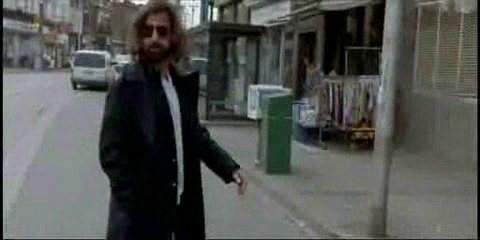}
  \includegraphics[width=0.1\linewidth]{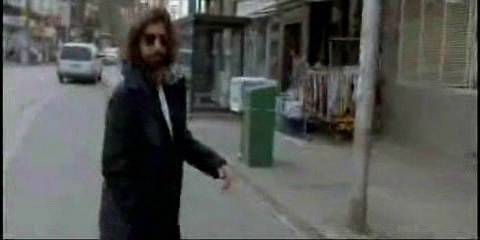} \\
  \includegraphics[width=0.1\linewidth]{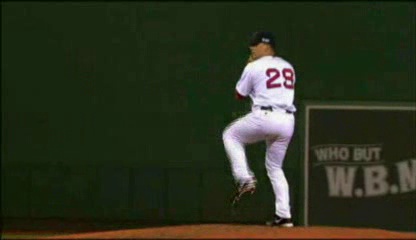}
  \includegraphics[width=0.1\linewidth]{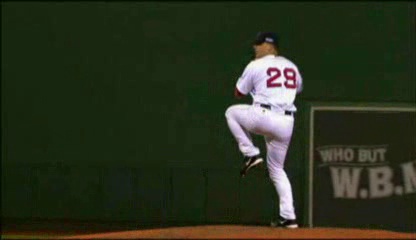}
  \includegraphics[width=0.1\linewidth]{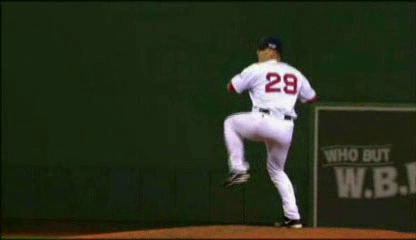}
  \includegraphics[width=0.1\linewidth]{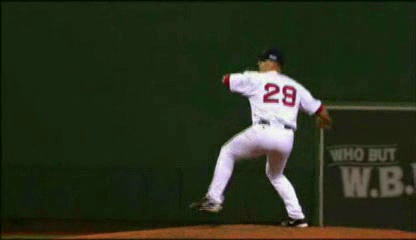}
  \includegraphics[width=0.1\linewidth]{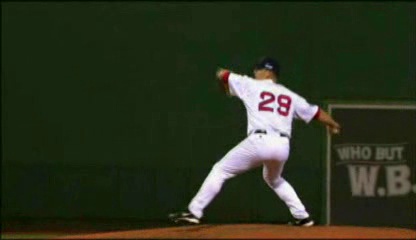}
  \includegraphics[width=0.1\linewidth]{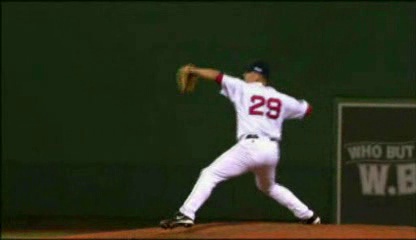}
  \includegraphics[width=0.1\linewidth]{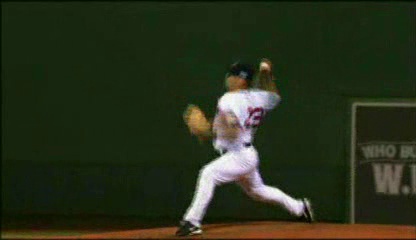}
  \includegraphics[width=0.1\linewidth]{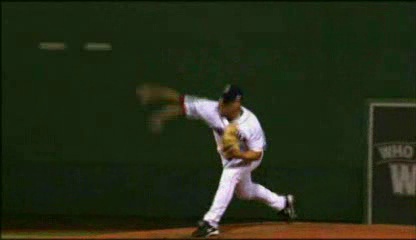}
  \includegraphics[width=0.1\linewidth]{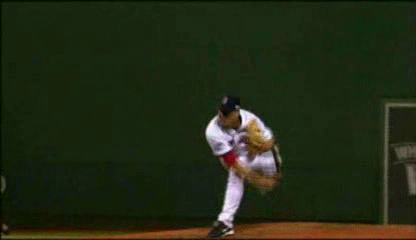} \\
  \includegraphics[width=0.1\linewidth]{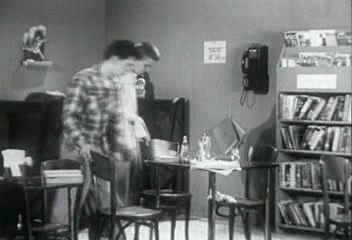}
  \includegraphics[width=0.1\linewidth]{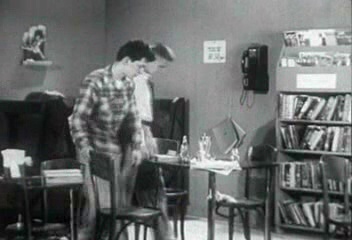}
  \includegraphics[width=0.1\linewidth]{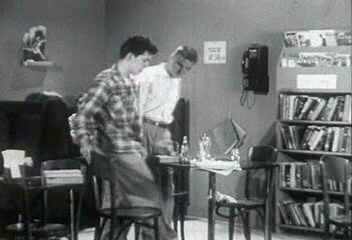}
  \includegraphics[width=0.1\linewidth]{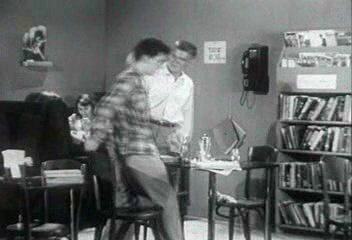}
  \includegraphics[width=0.1\linewidth]{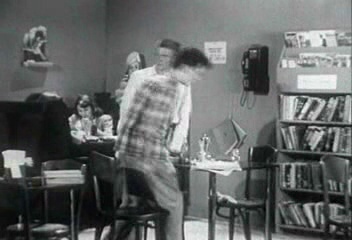}
  \includegraphics[width=0.1\linewidth]{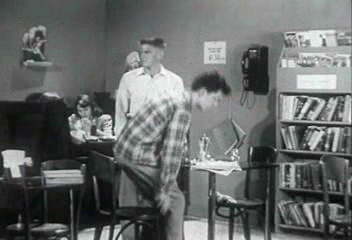}
  \includegraphics[width=0.1\linewidth]{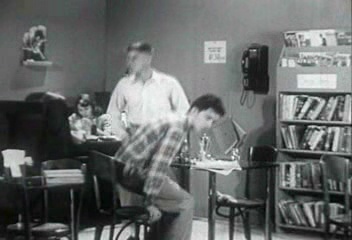}
  \includegraphics[width=0.1\linewidth]{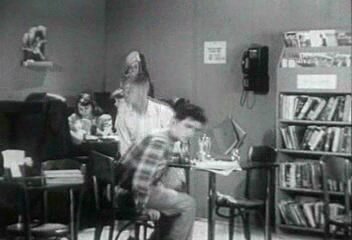}
  \includegraphics[width=0.1\linewidth]{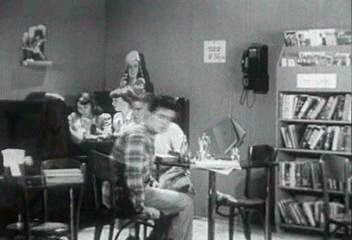} \\
  \includegraphics[width=0.1\linewidth]{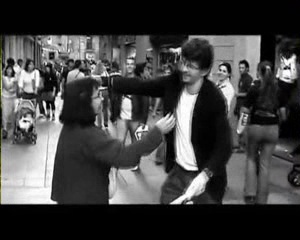}
  \includegraphics[width=0.1\linewidth]{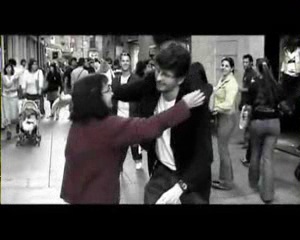}
  \includegraphics[width=0.1\linewidth]{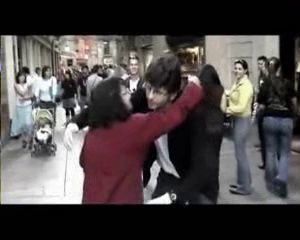}
  \includegraphics[width=0.1\linewidth]{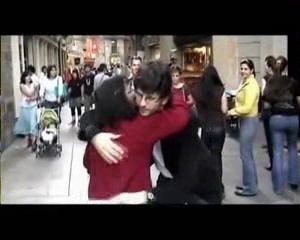}
  \includegraphics[width=0.1\linewidth]{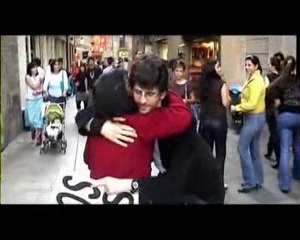}
  \includegraphics[width=0.1\linewidth]{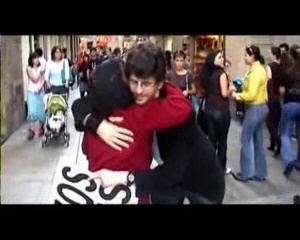}
  \includegraphics[width=0.1\linewidth]{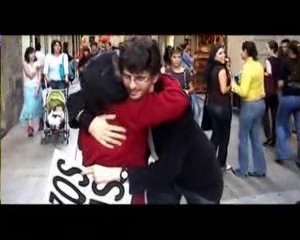}
  \includegraphics[width=0.1\linewidth]{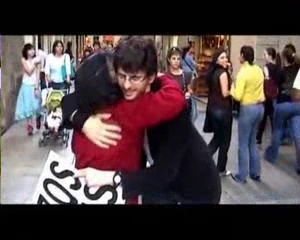} \\
  \includegraphics[width=0.1\linewidth]{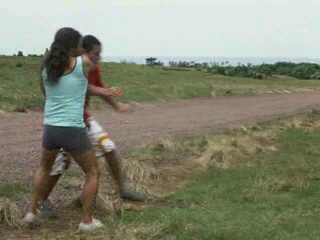}
  \includegraphics[width=0.1\linewidth]{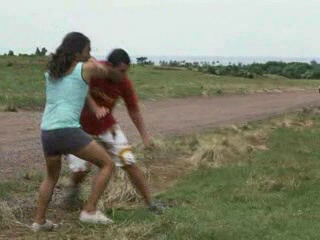}
  \includegraphics[width=0.1\linewidth]{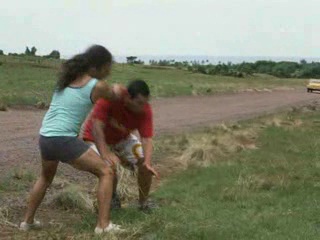}
  \includegraphics[width=0.1\linewidth]{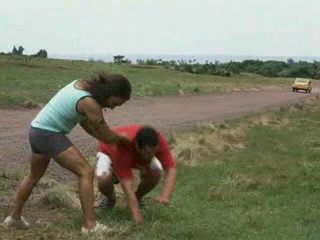}
  \includegraphics[width=0.1\linewidth]{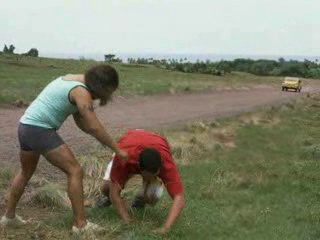}
  \includegraphics[width=0.1\linewidth]{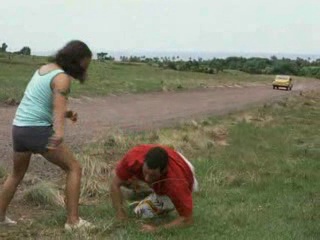}
  \includegraphics[width=0.1\linewidth]{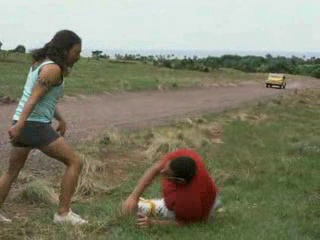}
  \includegraphics[width=0.1\linewidth]{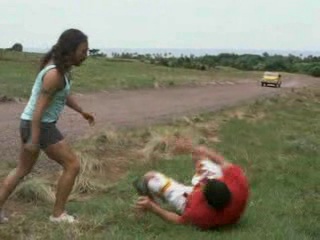}
  \includegraphics[width=0.1\linewidth]{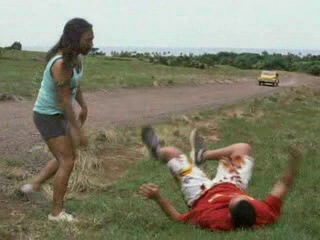} \\
  \includegraphics[width=0.1\linewidth]{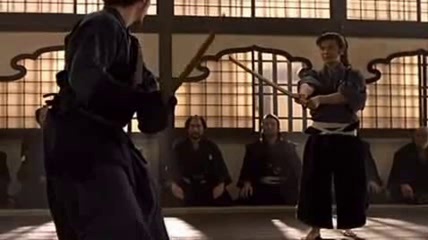}
  \includegraphics[width=0.1\linewidth]{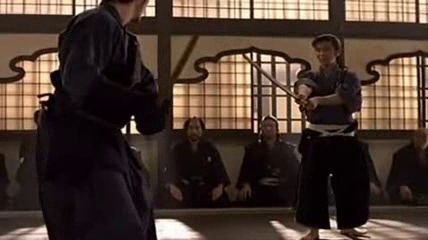}
  \includegraphics[width=0.1\linewidth]{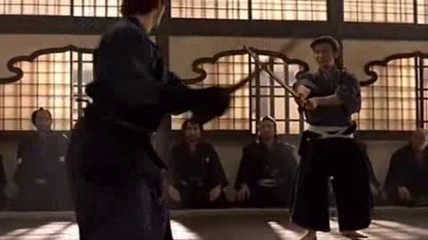}
  \includegraphics[width=0.1\linewidth]{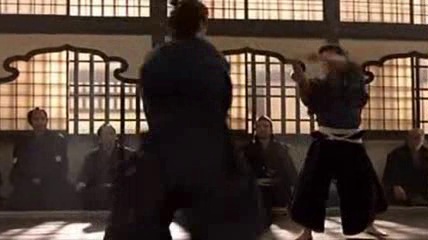}
  \includegraphics[width=0.1\linewidth]{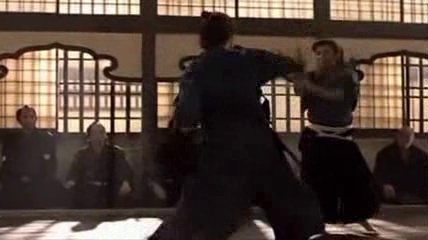}
  \includegraphics[width=0.1\linewidth]{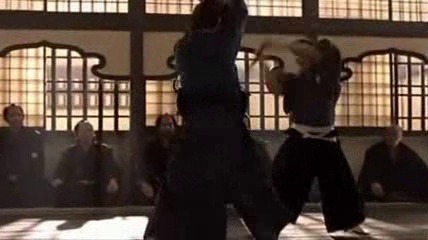}
  \includegraphics[width=0.1\linewidth]{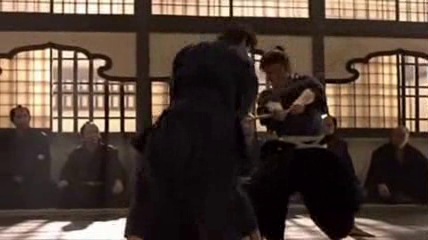}
  \includegraphics[width=0.1\linewidth]{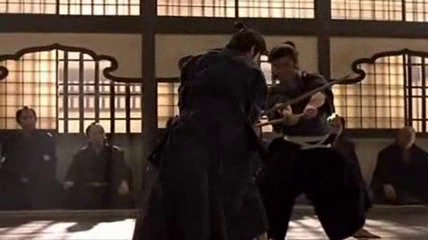}
  \includegraphics[width=0.1\linewidth]{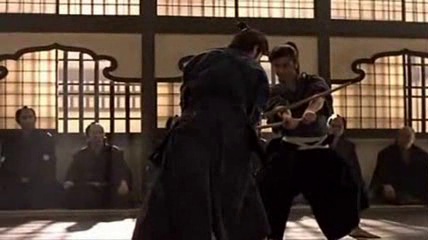} \\
  \includegraphics[width=0.1\linewidth]{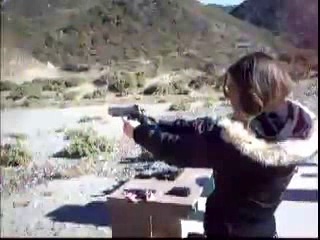}
  \includegraphics[width=0.1\linewidth]{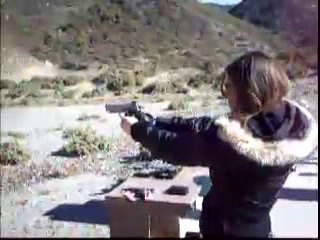}
  \includegraphics[width=0.1\linewidth]{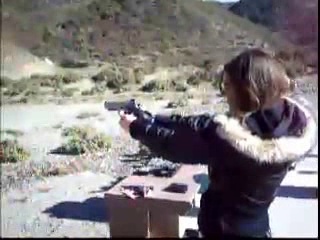}
  \includegraphics[width=0.1\linewidth]{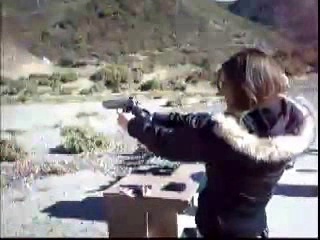}
  \includegraphics[width=0.1\linewidth]{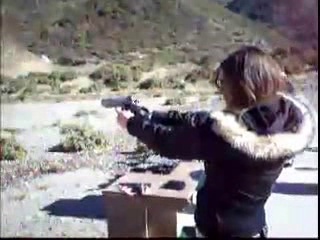}
  \includegraphics[width=0.1\linewidth]{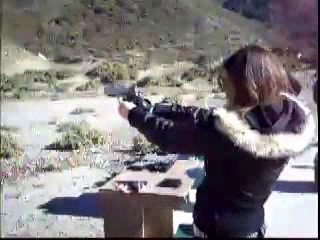}
  \includegraphics[width=0.1\linewidth]{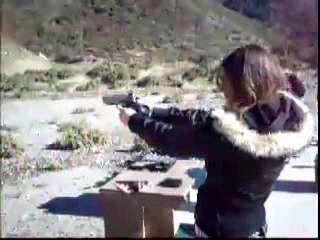}
  \includegraphics[width=0.1\linewidth]{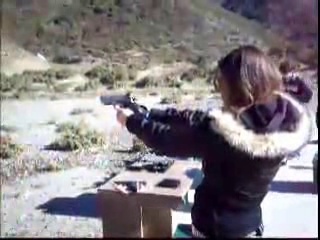}
  \includegraphics[width=0.1\linewidth]{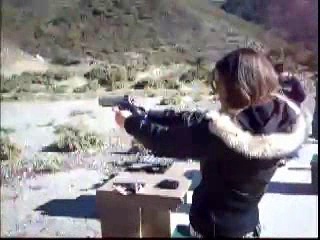} \\  
  \includegraphics[width=0.1\linewidth]{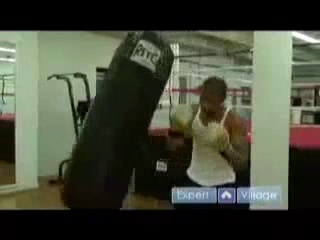}
  \includegraphics[width=0.1\linewidth]{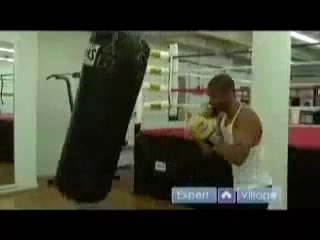}
  \includegraphics[width=0.1\linewidth]{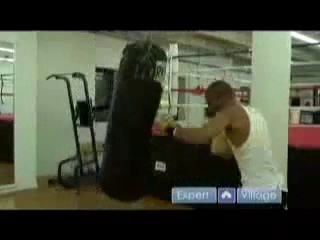}
  \includegraphics[width=0.1\linewidth]{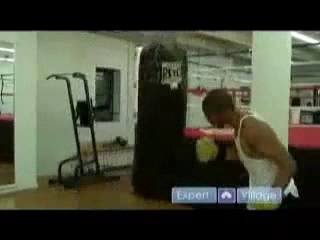}
  \includegraphics[width=0.1\linewidth]{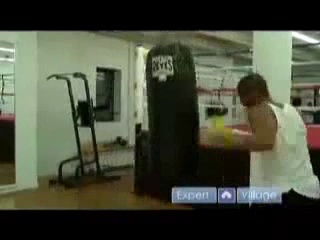}
  \includegraphics[width=0.1\linewidth]{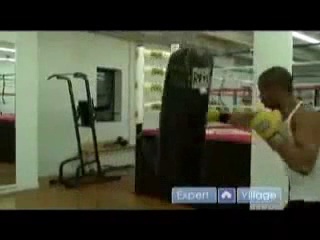}
  \includegraphics[width=0.1\linewidth]{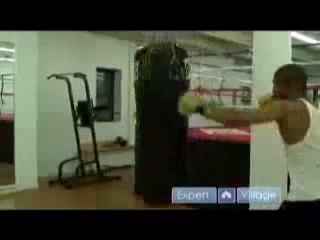}
  \includegraphics[width=0.1\linewidth]{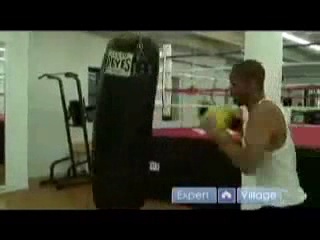}
  \includegraphics[width=0.1\linewidth]{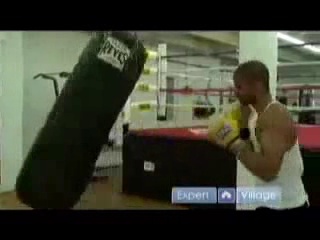}
  \end{center}
  \caption{\VIOset\ examples: wave, turn, throw, sit, hug, fall, sword, shoot, punch.}
\end{figure}

\newpage

\section{Videos for qualitative analyses}
\label{app:qual_examples}

\begin{figure}[h]
\begin{center}
  \includegraphics[width=0.24\linewidth]{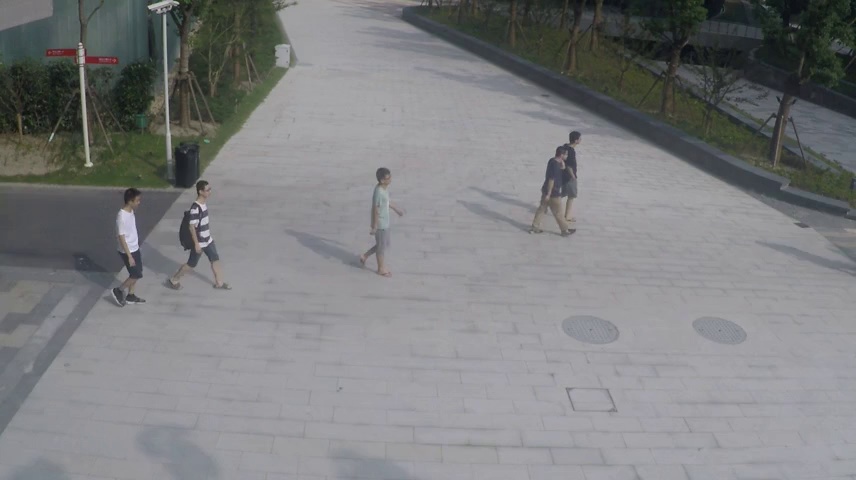}
  \includegraphics[width=0.24\linewidth]{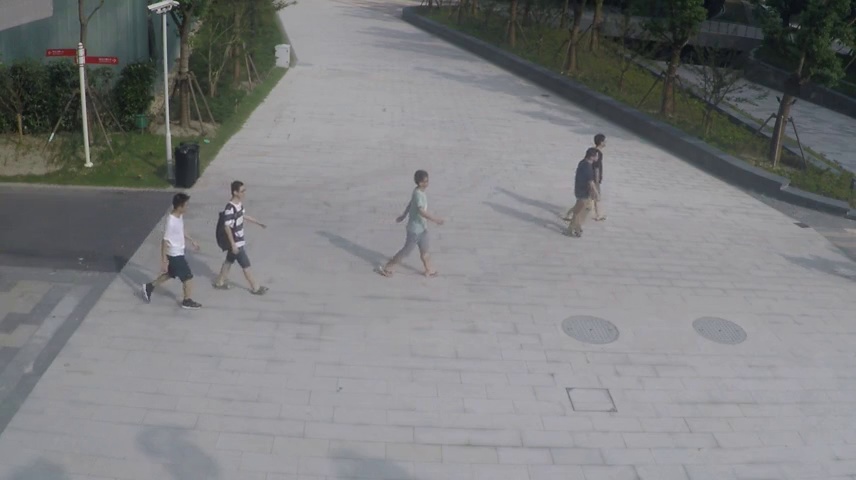}
  \includegraphics[width=0.24\linewidth]{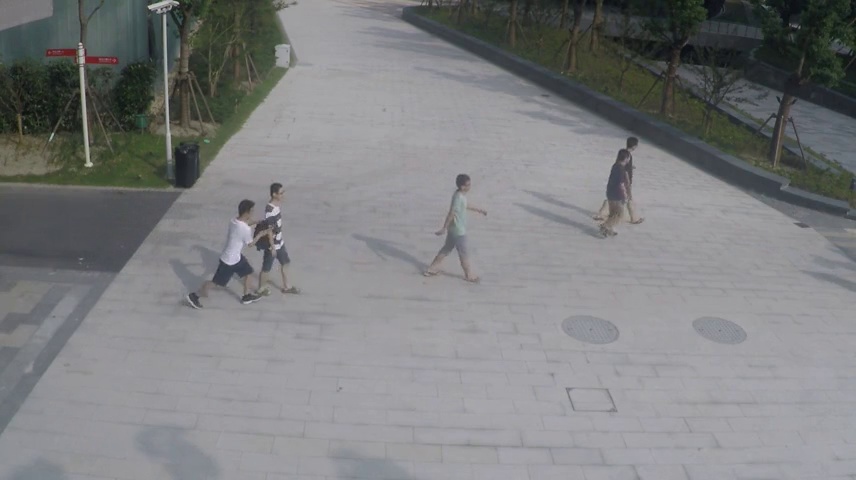}
  \includegraphics[width=0.24\linewidth]{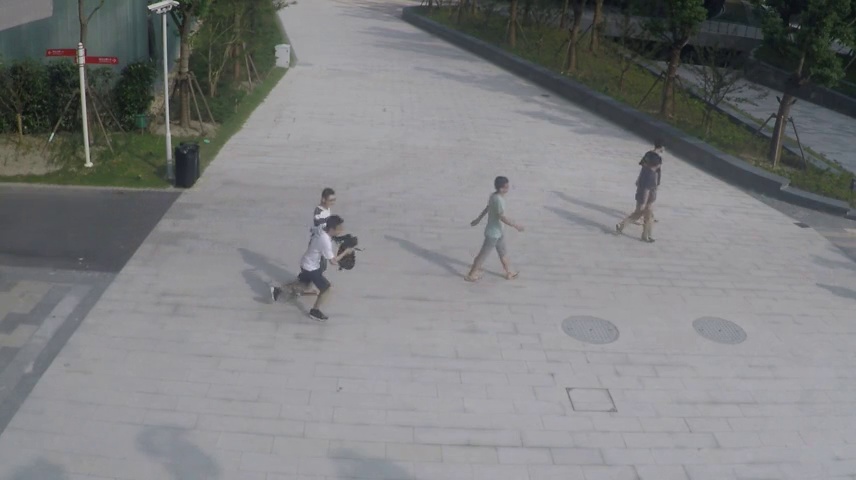} \\
  \includegraphics[width=0.24\linewidth]{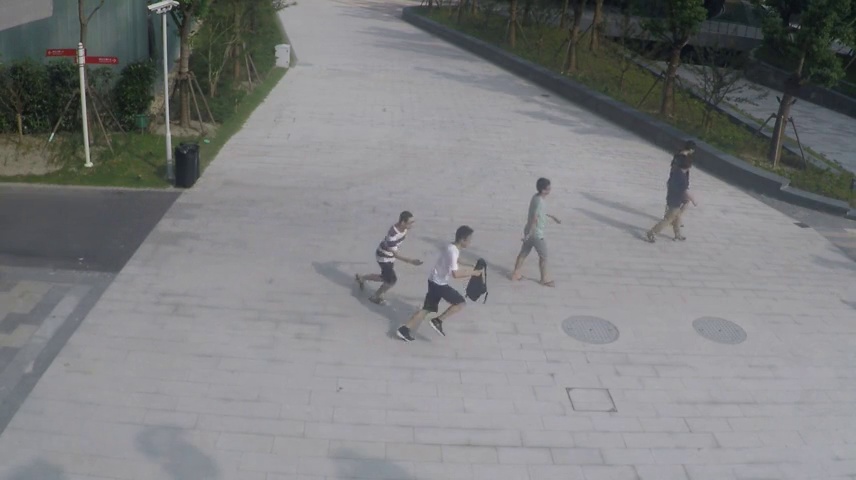}
  \includegraphics[width=0.24\linewidth]{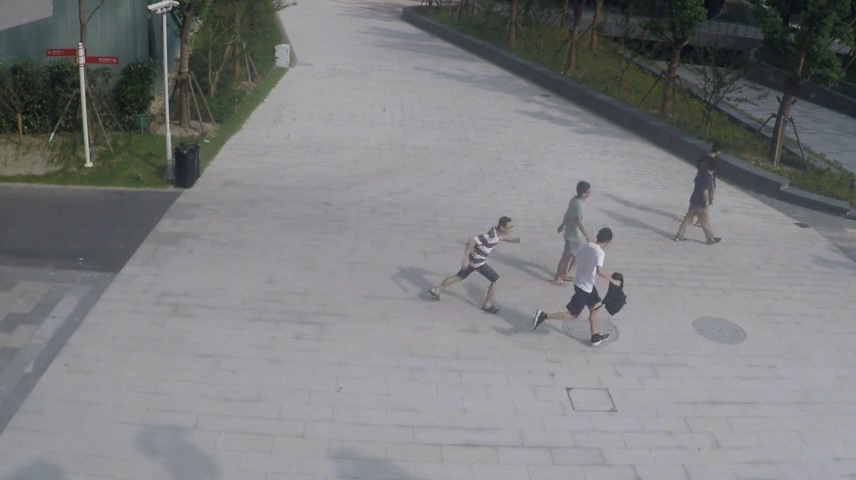}
  \includegraphics[width=0.24\linewidth]{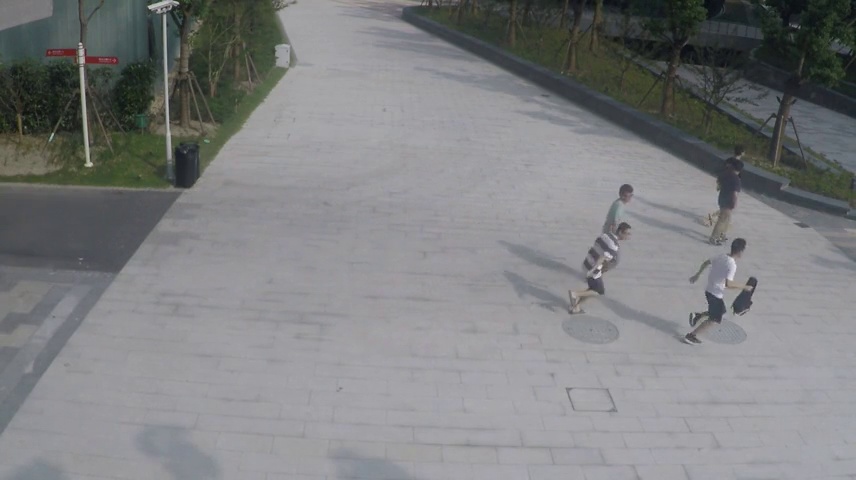}
  \includegraphics[width=0.24\linewidth]{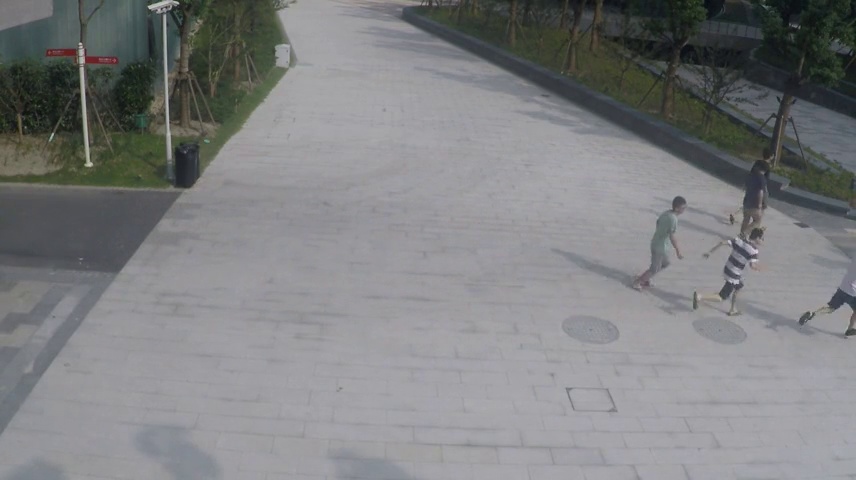} \\
  \vspace{0.5cm}
  \includegraphics[width=0.24\linewidth]{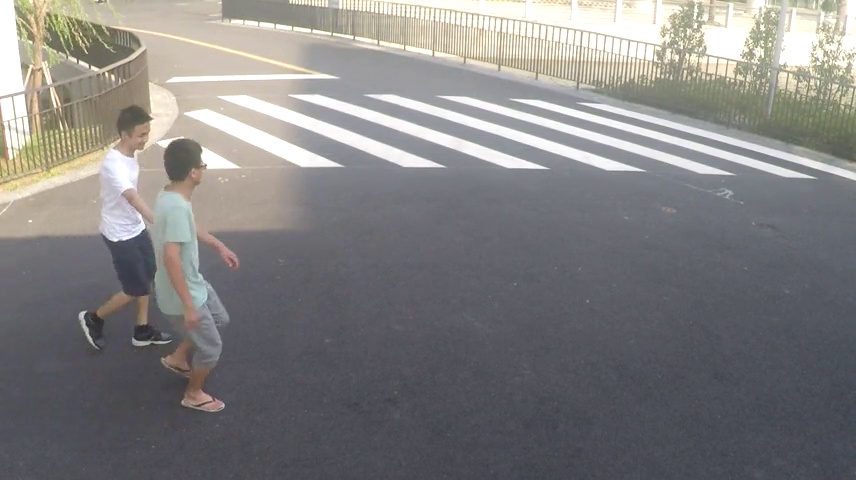}
  \includegraphics[width=0.24\linewidth]{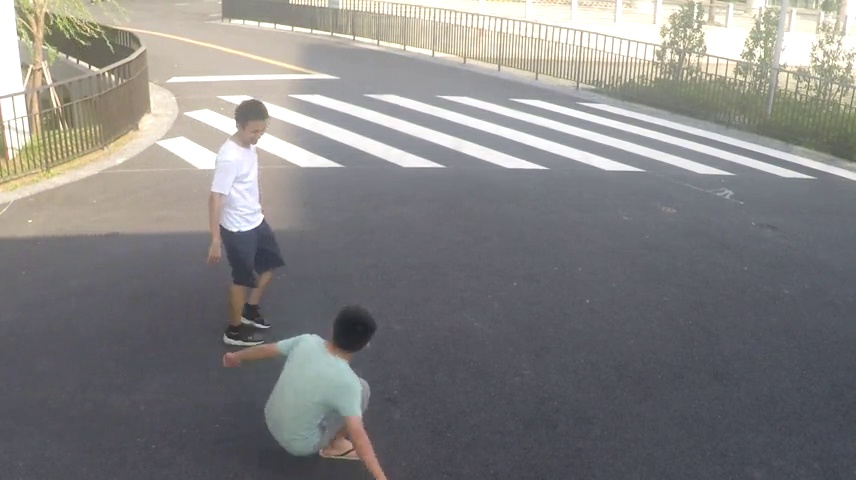}
  \includegraphics[width=0.24\linewidth]{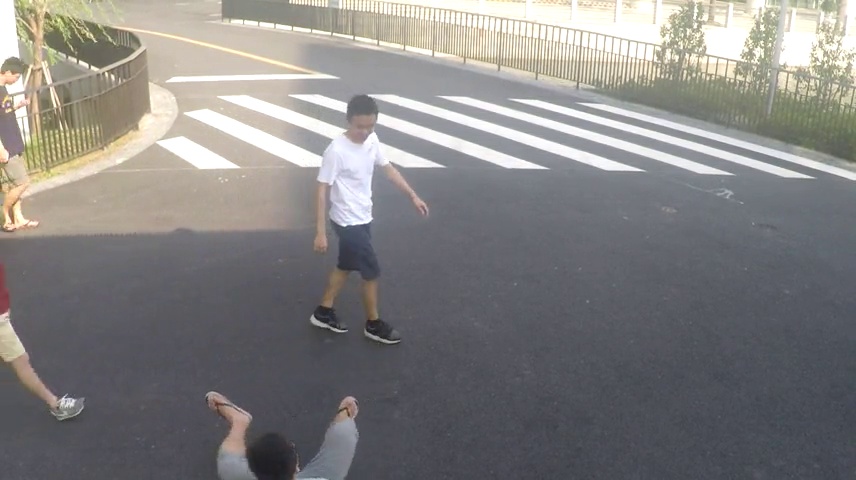}
  \includegraphics[width=0.24\linewidth]{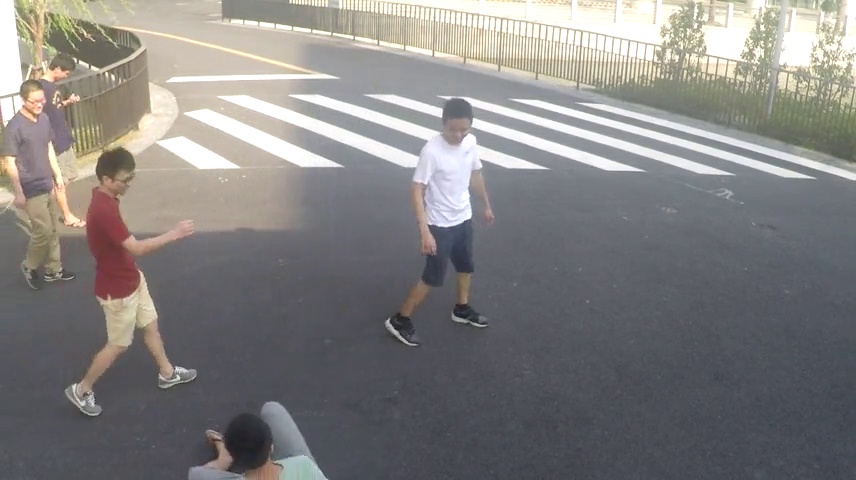} \\
  \includegraphics[width=0.24\linewidth]{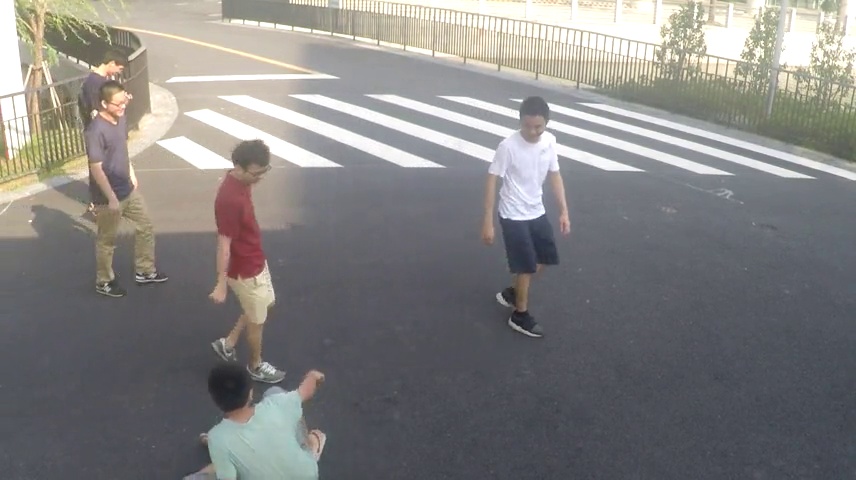} 
  \includegraphics[width=0.24\linewidth]{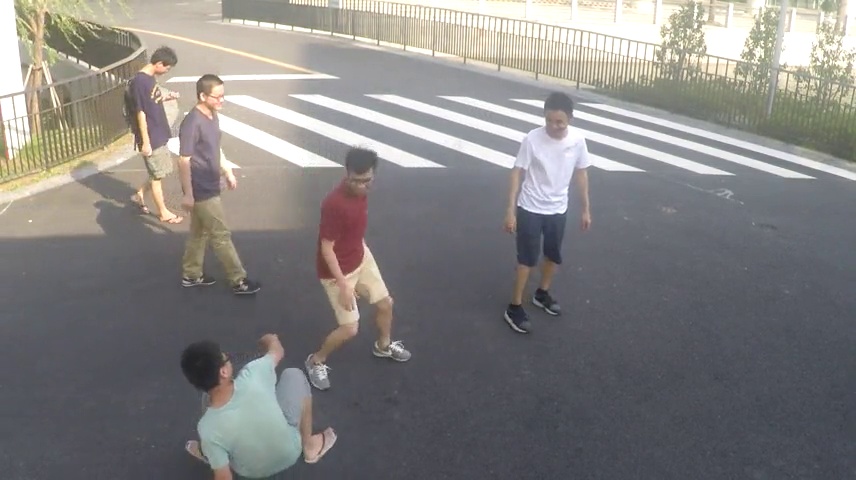}
  \includegraphics[width=0.24\linewidth]{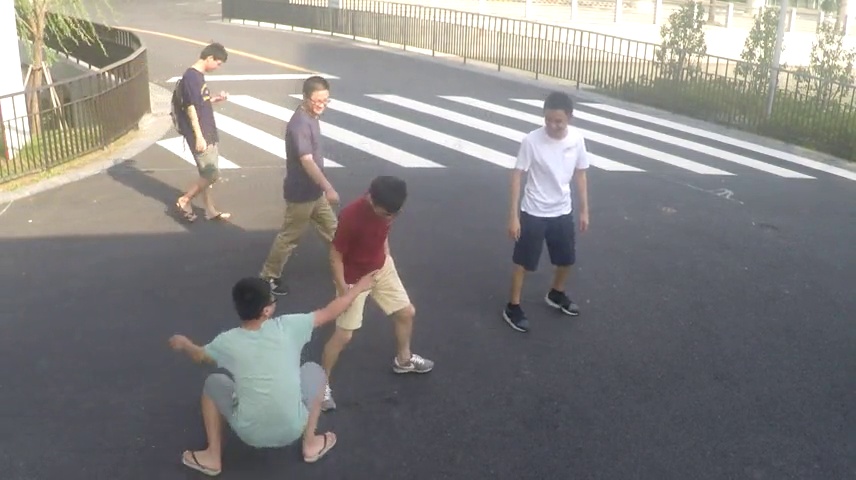}
  \includegraphics[width=0.24\linewidth]{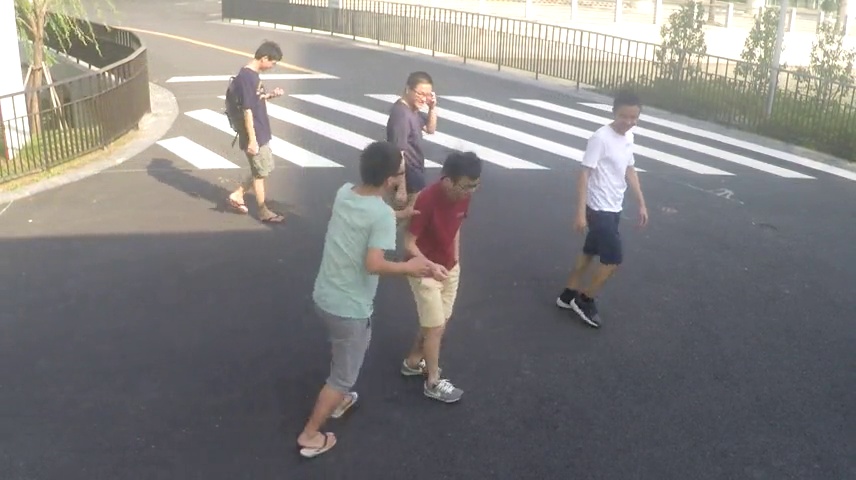} \\
  \vspace{0.5cm}
  \includegraphics[width=0.24\linewidth]{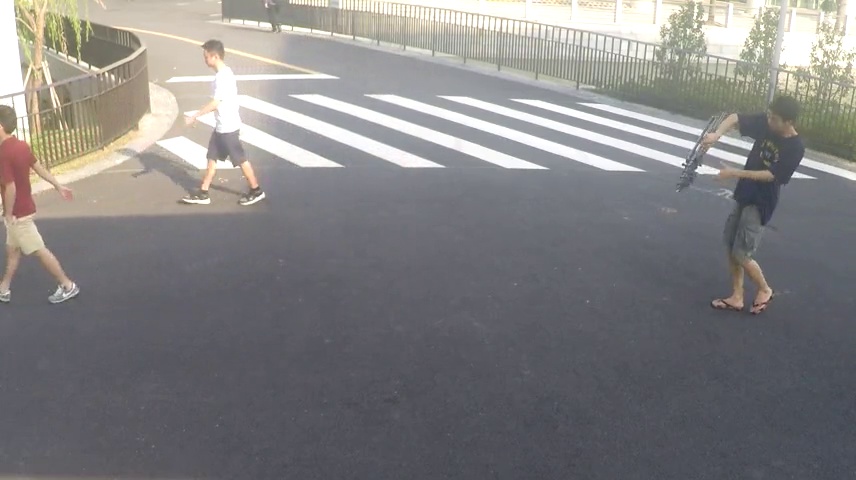}
  \includegraphics[width=0.24\linewidth]{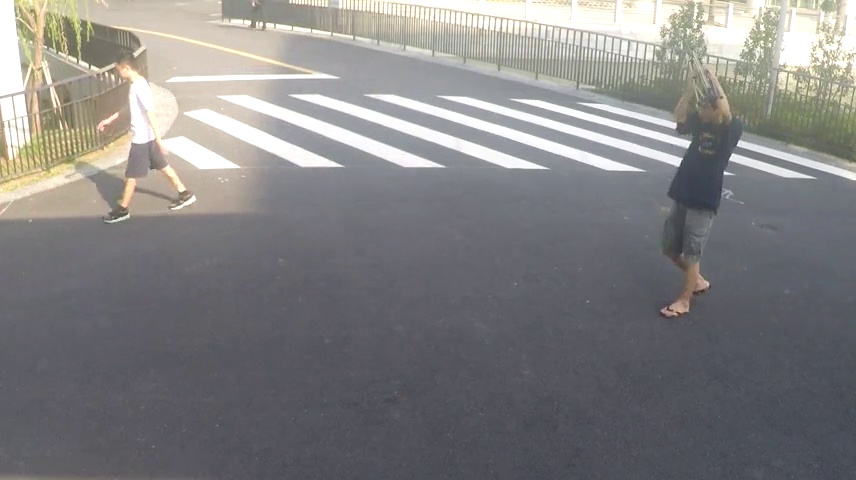}
  \includegraphics[width=0.24\linewidth]{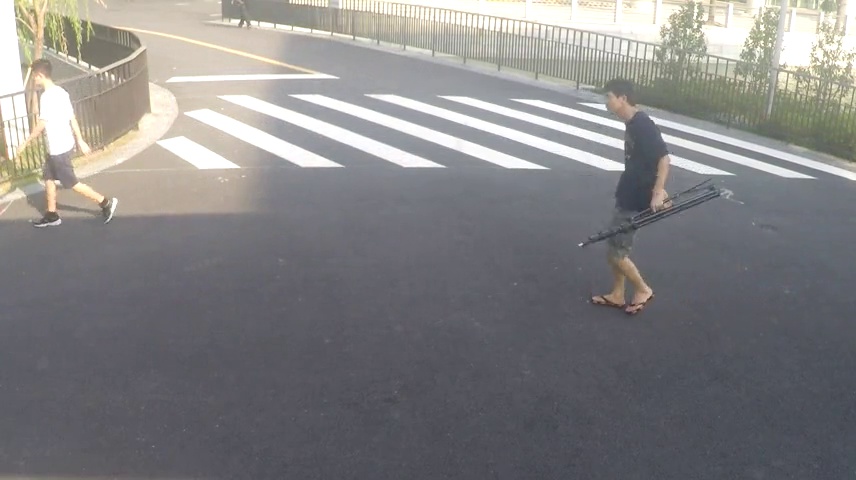}
  \includegraphics[width=0.24\linewidth]{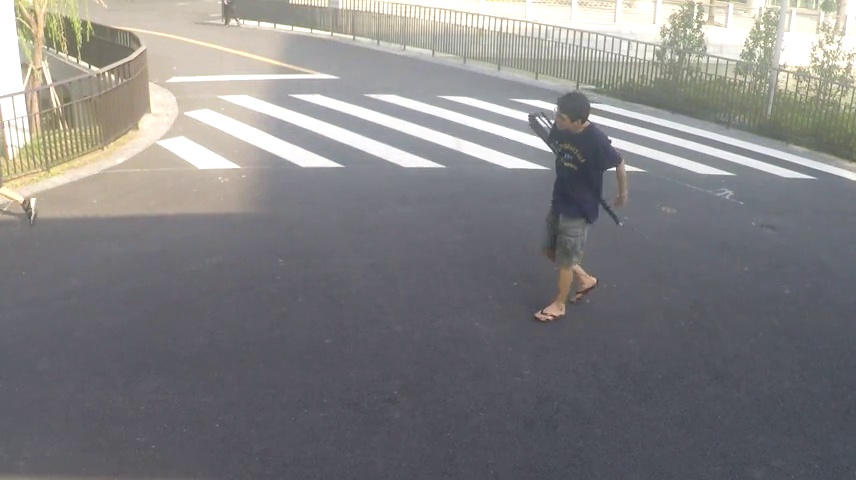} \\
  \includegraphics[width=0.24\linewidth]{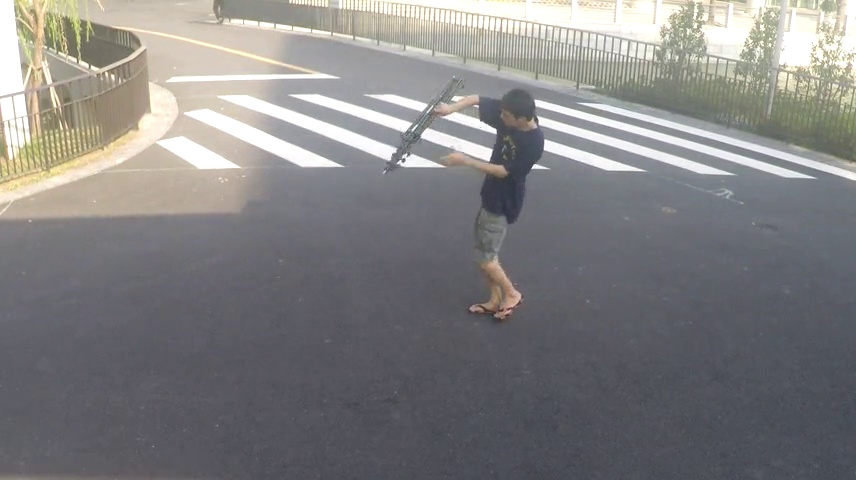}
  \includegraphics[width=0.24\linewidth]{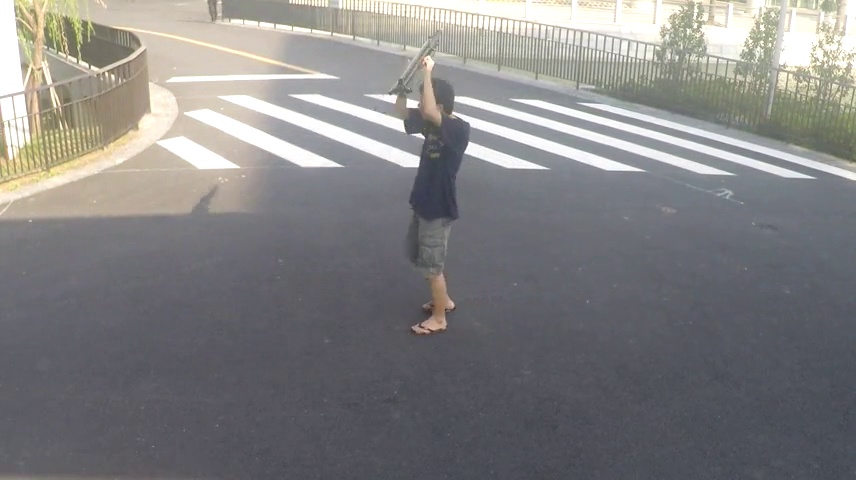}
  \includegraphics[width=0.24\linewidth]{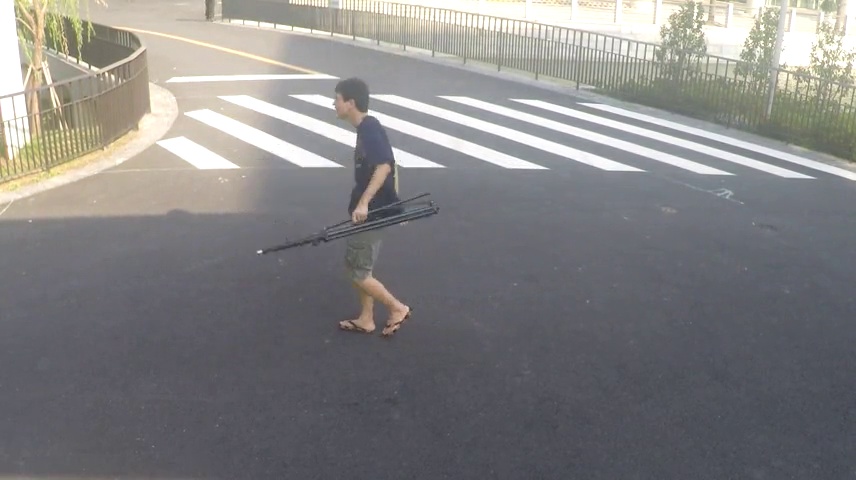}
  \includegraphics[width=0.24\linewidth]{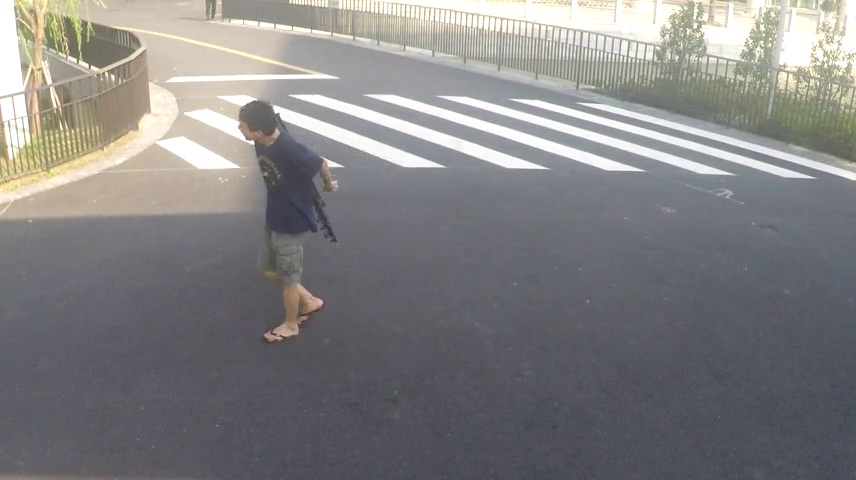} \\
  \vspace{0.5cm}
  \includegraphics[width=0.24\linewidth]{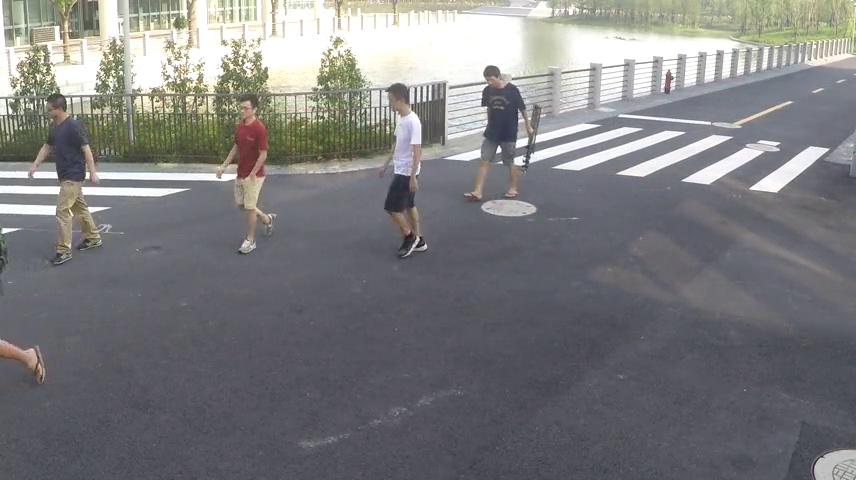}
  \includegraphics[width=0.24\linewidth]{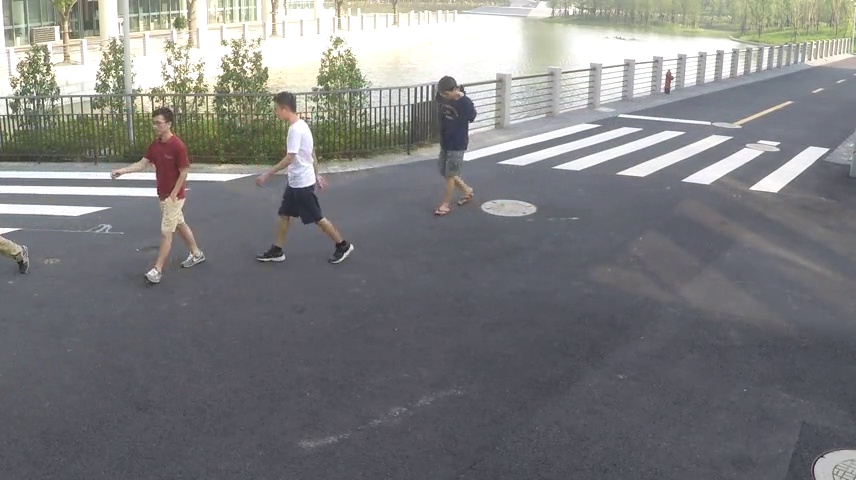}
  \includegraphics[width=0.24\linewidth]{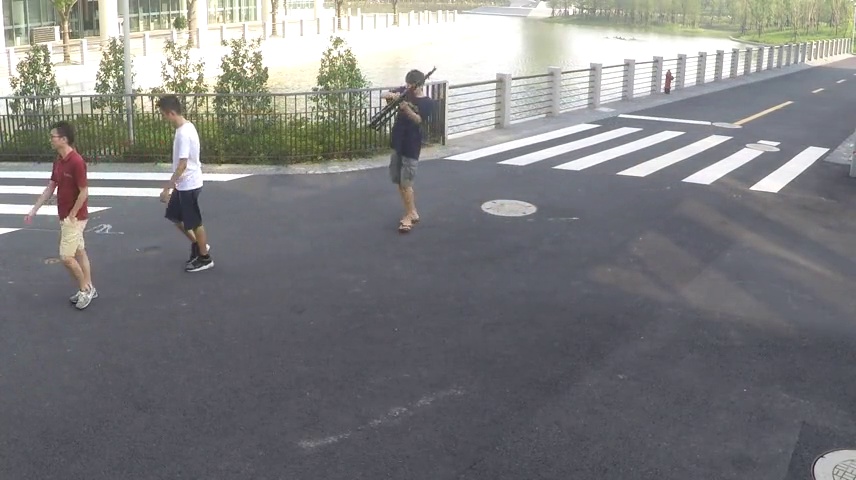}
  \includegraphics[width=0.24\linewidth]{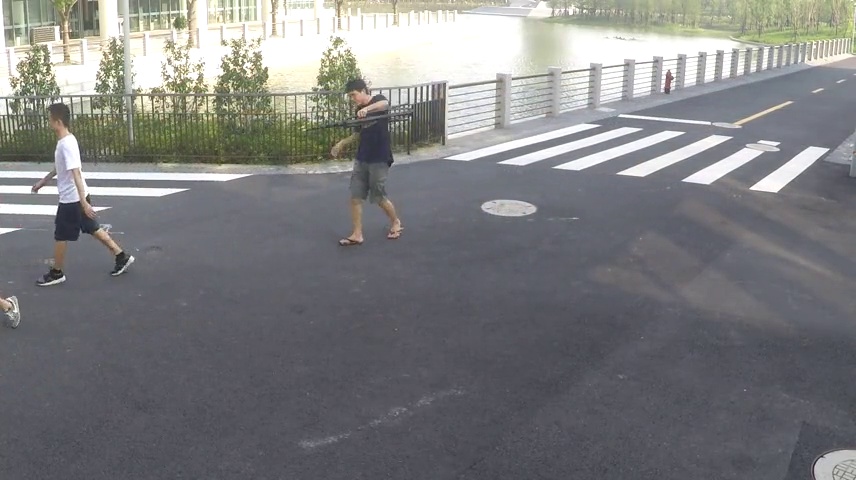} \\
  \includegraphics[width=0.24\linewidth]{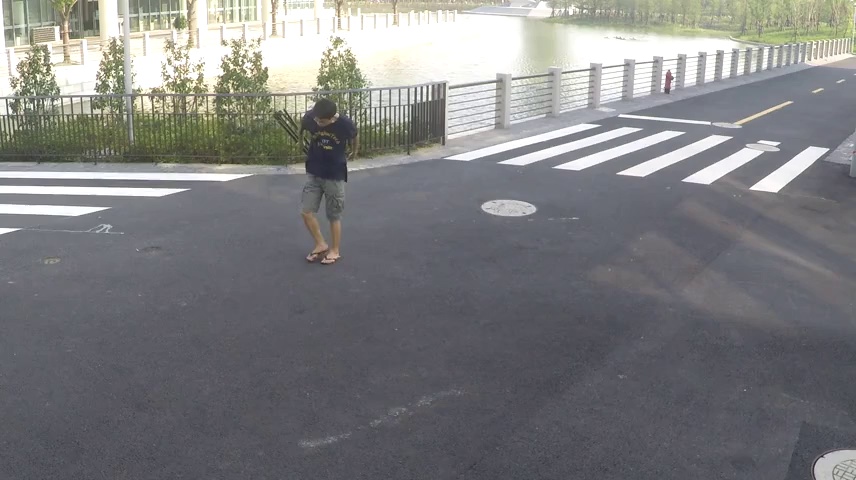}
  \includegraphics[width=0.24\linewidth]{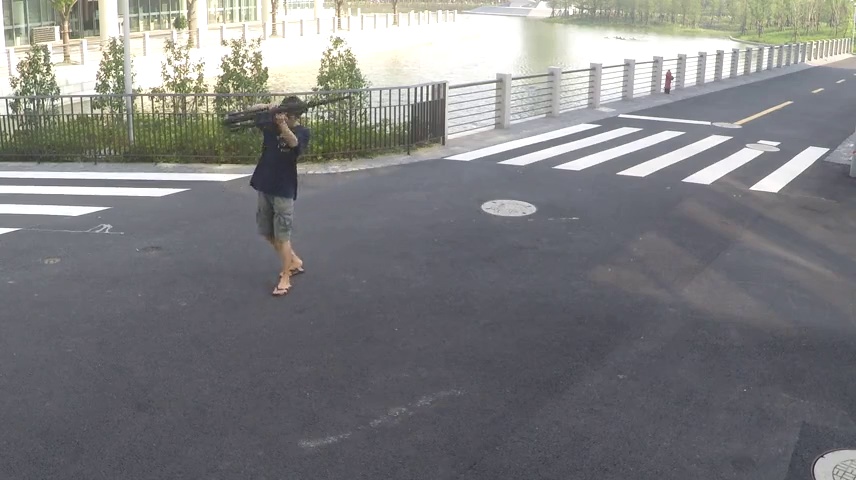}
  \includegraphics[width=0.24\linewidth]{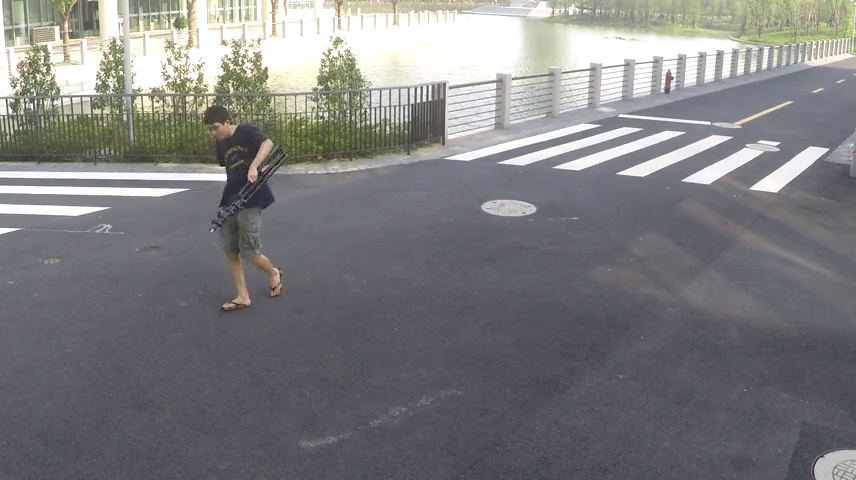}
  \includegraphics[width=0.24\linewidth]{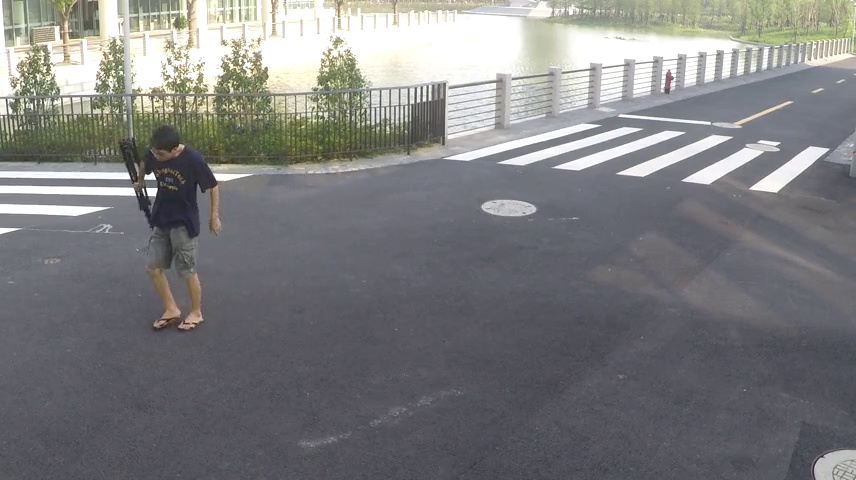}
  \end{center}
  \caption{The anomalies from 01\_0028, 03\_0032, 03\_0039, 07\_0008 (top to bottom, respectively) videos from ShanghaiTech Campus dataset.}
\end{figure}

\newpage

\end{document}